\documentclass[10pt,twocolumn,letterpaper]{article}

\usepackage{iccv}
\usepackage{times}
\usepackage{epsfig}
\usepackage{graphicx}
\usepackage{amsmath}
\usepackage{amssymb}
\usepackage{booktabs}
\usepackage{enumitem}
\usepackage{appendix}
\usepackage{float}

\usepackage[pagebackref=true,breaklinks=true,letterpaper=true,colorlinks,bookmarks=false]{hyperref}

\iccvfinalcopy 

\def\iccvPaperID{4911} 

\ificcvfinal\fi

\begin{document}

\title{Benchmarking the Reliability of Post-training Quantization: \\a Particular Focus on Worst-case Performance}

\author{
Zhihang Yuan$^{1,\star}$
~~~
Jiawei Liu$^{2,\star,\diamond}$
~~~
Jiaxiang Wu$^{3}$
~~~
Dawei Yang$^{1}$
~~~
Qiang Wu$^{1}$\\
Guangyu Sun$^{4}$
~~~
Wenyu Liu $^{2}$
~~~
Xinggang Wang$^{2,\dagger}$
~~~
Bingzhe Wu$^{3,\dagger}$\\
$^1$ Houmo AI ~~~
$^2$ School of EIC, Huazhong University of Science \& Technology \\
$^3$ Tencent AI Lab ~~~$^4$ Peking University}

\maketitle
\ificcvfinal\thispagestyle{empty}\fi

\let\thefootnote\relax\footnotetext{$^\star$ Equal contribution (\texttt{hahnyuan@gmail.com, jiaweiliu@\\hust.edu.cn}). $^\diamond$ This work was done when Jiawei Liu was an intern at Houmo AI. $^\dagger$ Corresponding authors (\texttt{wubingzheagent@gmail.com, xgwang@hust.edu.cn}).}

\begin{abstract}
   Post-training quantization (PTQ) is a popular method for compressing deep neural networks (DNNs) without modifying their original architecture or training procedures. 
   Despite its effectiveness and convenience, the reliability of PTQ methods in the presence of some extrem cases such as distribution shift and data noise remains largely unexplored. This paper first investigates this problem on various commonly-used PTQ methods. We aim to answer several research questions related to the influence of calibration set distribution variations, calibration paradigm selection, and data augmentation or sampling strategies on PTQ reliability. A systematic evaluation process is conducted across a wide range of tasks and commonly-used PTQ paradigms. The results show that most existing PTQ methods are not reliable enough in term of the worst-case group performance, highlighting the need for more robust methods. Our findings provide insights for developing PTQ methods that can effectively handle distribution shift scenarios and enable the deployment of quantized DNNs in real-world applications.
\end{abstract}

\section{Introduction}



\begin{figure}[tb] 
    \centering
    \includegraphics[width=0.48\textwidth]{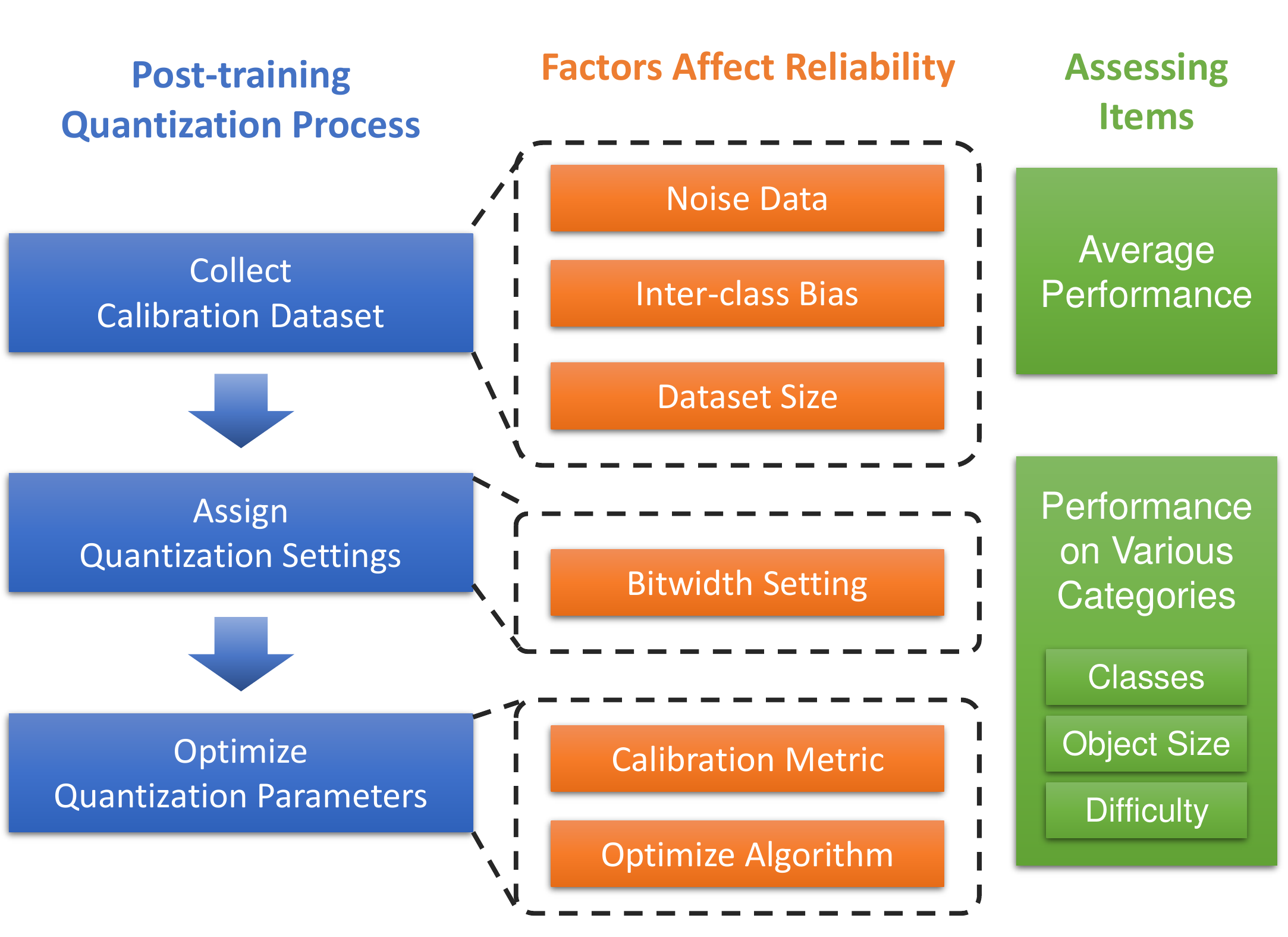} 
    \caption{Overview of the framework for assessing reliability of PTQ.}
    \label{im_overview}
\end{figure}

Modern deep neural networks (DNNs) are widely used in various domains ranging from daily-life applications such as image classification and machine translation to risk-sensitive areas such as autonomous driving \cite{muhammad2020deeplearning_autodrive} and finance \cite{zhang2021application_finance}. The remarkable performance of DNNs always depends on huge amounts of network parameters, which further brings expensive computational and memory costs. Prior research aims to compress DNN's parameters from different directions. Among these directions, post-training quantization (PTQ)
is typically identified as one of the most practical way, which offers the advantage of compressing DNNs without modifying the original training procedure, model structures, and parameters, making them highly desirable for practical applications.

At a high-level, PTQ is a technique for compressing neural networks by reducing the number of bits used to represent the weights and activations of the network. 
The calibration process of PTQ determines the optimal quantization parameters for a given neural network. 
The calibration dataset is small number of input samples, which is used to estimate the statistical properties of the activation values that the neural network produces during inference.
These statistical properties are then used to determine the optimal quantization parameters that minimize the quantization error.
Previous efforts have been made to improve PTQ from different dimensions such as proposing better calibration metrics~\cite{yuan2022ptq4vit,liu2022pdquant} and devising new optimizing algorithms~\cite{nagel2020adaround,li2021brecq}, especially in the low-bit regime. 
To date, state-of-the-art method can achieve nearly lossless accuracy on the image classification task in the 4-bit setting~\cite{wang2022leveraging}. 

While the PTQ technique has been found to be effective and convenient, they are often based on the assumption that the calibration, and test sets share the same underlying distribution, a condition known as the "close-environment" assumption. 
In real-world open environment, however, this assumption is typically impractical due to distribution shifts between the calibration and test distributions. 
A natural question arises: \textit{Is current PTQ method reliable enough when facing extreme test samples such as worst-case-category or out-of-distribution samples?}. 
Answering this question is crucial for deploying the quantized DNNs to real-world applications. 
Despite its importance, this question remains largely unexplored currently in the research community.

In this paper, the reliability of various commonly-used PTQ methods are deeply investigated and comprehensively evaluated for the first time. 
In contrast to most of previous work that only focus on the average performance, we aim to evaluate PTQ's performance on the worst-case subgroups in the test distribution. To this end, we first perform extensive experiments and observe that some specific test categories suffer significant performance drop after post-training quantization.


In practical applications, we further found that PTQ's performance is influenced by numerous factors, including the distribution of the calibration set, PTQ setting, and calibration metrics. 
We delve deeper into the influence of these factors and their collective impact on the reliability of previous methods in the face of the worst test cases such as subpopulation shift. 
Specifically, we aim to answer the following research questions:
\begin{itemize}[itemsep=0pt]
    \item How does the variation in the distribution of the calibration data affect the performance of quantized network?
    \item How do different quantization settings affect the performance among different categories?
    \item How do different PTQ methods affect the performance among different categories?
\end{itemize}
To answer these questions, this paper introduces a framework designed to systematically evaluate the impact of PTQ from a reliability perspective. 
Using this framework, we conducted experiments spanning a broad range of tasks and commonly-used PTQ paradigms.
For each of the research questions, we present quantitative results, along with a thorough analysis of the underlying factors.
The primary observations derived from our experiments are summarized as follows:
\begin{itemize}[itemsep=0pt]
    \item Significant accuracy drop of individual categories  is observed  on the quantized model, which indicates the reliability issue of current PTQ methods. (Sec~\ref{sec_evaluation_method})
    \item The average prediction accuracy remains resilient to variations in the composition of the calibration dataset, in the presence of noise and intra-class bias. However, the prediction accuracy for individual categories displays sensitivity to these factors.(Sec~\ref{sec_calibration_dataset})
    \item Reducing the bit-width has the potential to lead to substantial degradation in specific categories. (Sec~\ref{sec_quantization_settings})
    \item Certain optimization algorithms, such as gradient-based PTQ, exhibit comparatively diminished reliability, despite attaining superior average predictive accuracy. (Sec~\ref{sec_optimization})
\end{itemize}
From the above observations, it is apparent that most of the existing mainstream methods are not reliable enough.
In certain cases, the quantized model suffers significant accuracy drop on some categories or groups (e.g., small object in the object detection), which is unacceptable for risk-sensitive scenarios.
We indicate it is necessary to devise reliable and more robust PTQ methods that can effectively handle distribution shift scenarios.
As an additional contribution, we provide a benchmark for PTQ reliability that encompasses various tasks, PTQ methods, and network architectures. 
We will further release this benchmark to facilitate future analyses of PTQ reliability and contribute to the advancement of PTQ methods.



\section{Related Work}

\subsection{Quantization}

Quantization is one of the most effective methods to compress neural networks and to accelerate the inference of networks~\cite{deng2020model_compress_survey,nagel2021white,yuan2022ptq4vit}.
Quantizing a network means reducing the precision of the weights and activations of the network in order to make it more computationally efficient while maintaining a similar level of prediction accuracy. 
The quantized weights and activations are typically stored as integers (such as INT8 and INT4), which reduces the amount of memory required to store them.
Lower precision calculations can be performed faster than higher precision calculations, resulting in faster inference times and reduced energy in real-time applications.
There are two main types of quantization used for neural networks: Post-Training Quantization (PTQ) and Quantization Aware Training (QAT).

Post-Training Quantization (PTQ) is a quantization technique that is applied to a pre-trained neural network~\cite{migacz2017tensorrt,banner2019ptq4bit_rapid_deploy,choukroun2019lowbit_for_efficient_inference}. 
It involves quantizing the weights and activations of the network after training is complete. 
Quantization Aware Training (QAT) is a technique in which the quantization process is integrated into the training process itself~\cite{krishnamoorthi2018quantizing_whitepaper,choi2018pact,gong2019dsq,esser2020lsq}. 
During QAT, the network is trained using a combination of full-precision and lower-precision weights and activations, which helps it to learn to be more robust to the effects of quantization.
PTQ is generally simpler to implement than QAT, as it involves quantizing a pre-trained network without any additional training. 
This can be useful in situations where re-training the network with QAT would be time-consuming or impractical~\cite{nagel2021white} (e.g., when the training dataset is not available).

Some previous work has explored the stability of quantization and the influence of calibration dataset.
\cite{hubara2021ptq_with_small_calibraiton_set} explore the problems of using small calibration dataset in quantization. 
PD-Quant~\cite{liu2022pdquant} identifies a disparity between the distribution of calibration activations and their corresponding real activations, and proposes a technique for adjusting the calibration activations accordingly.
SelectQ~\cite{zhang2022selectq} shows that randomly selecting data for calibration in PTQ can result in performance instability and degradation due to activation distribution mismatch.
Although prior research has explored the impact of the calibration dataset on the performance of quantized neural networks, these investigations have only focused on a limited range of factors and their effects on the stability of quantization. 
In this paper, we present a framework that enables us to evaluate the reliability of PTQ and conduct a comprehensive analysis of the impact of various factors on the reliability of quantization.


\subsection{Reliability of Neural Network}

The reliability of deep models has become an increasingly important topic in recent years, particularly as these models are increasingly being used in critical applications such as healthcare, autonomous vehicles, and financial systems. The reliability of deep models is often measured through different dimensions, some commonly used dimensions include: (1) model performance in various situations (including performance on the existing worst categories, noise samples, and out-of-distribution samples, etc.), (2) model robustness against test-time attacks such as adversarial attacks, and (3) the quality of the model's confidence. This paper mainly focus on the first dimension and leaves other two as the future work. Therefore, here we only give a brief introduction of related works in the first dimension.

\noindent \textbf{Model reliability in the worst case:} For the first dimension, various studies have been conducted to evaluate and improve the reliability of models against worst-case scenarios involving distribution shift and data noise, among other factors\cite{OOD_survey, label_noise_survey}. In this context, the evaluation metric typically used is the worst-case accuracy among existing test categories or out-of-distribution (OOD) samples\cite{OOD_survey, dro_survey}. It has been observed that models trained conventionally with the assumption of independent and identically distributed (IID) data fail to generalize in real-world testing environments with challenges such as hardness, noise, or OOD samples\cite{co-teaching, duchi_dro}. To enhance model reliability in such settings, two common approaches have been proposed: (1) explicit employment of robust training paradigms, such as distributionally robust optimization (DRO) and learning with noisy labels (LNL), to enhance model robustness; and (2) implicit data augmentation techniques, such as Mixup\cite{zhang2018mixup}. Further literature on this subject can be found in prior surveys\cite{OOD_survey, label_noise_survey, dro_survey}. Despite the efforts made by the community, the model reliability of PTQ methods in such cases remains largely unexplored.

\section{Exploring Reliability of PTQ}

In this section, we will initially introduce the generic workflow of the Post-Training Quantization (PTQ). 
Subsequently, we will define the PTQ reliability and propose an approach to examine the reliability of PTQ.

\subsection{PTQ Workflow}

PTQ is an effective method for compressing neural networks and accelerating their inference.
The workflow of PTQ involves three key steps, namely collecting a calibration dataset, assigning quantization settings, and optimizing quantization parameters. 
While it is worth noting that not all prior work can be neatly subsumed under these three steps, these stages are consistently found in the majority of PTQ methodologies.

\textbf{Collecting Calibration Dataset}: PTQ requires a calibration dataset, which is a number of input samples, to compute the activations of each layer, $\{X^1,...,X^L\}$.
The majority of academic papers typically obtain their calibration dataset through random sampling from the training dataset. 
In contrast, industrial applications involve the user collecting a specific amount of real-world data to serve as the calibration dataset.


\textbf{Assigning Quantization Settings}: The next step is to choose the quantization settings, which specify the bit-width and the quantization function. 
The quantization bit-width refers to the number of bits used to represent a numerical value in a quantized representation.
The quantization function determines how the continuous values of weights and activations are mapped to discrete values, such as uniform quantization~\cite{krishnamoorthi2018quantizing_whitepaper} and non-uniform quantization~\cite{miyashita2016conv_logarithmic_data,yuahngli2020apot}. 
Taking the symmetric uniform quantization function as an example, float value $x$ is quantized to $k$ bits integer $x_q$:
\begin{equation}
    x_q = clamp(round(\frac{x}{s}),-2^{k-1},2^{k-1}-1),
\label{quantization_function}
\end{equation}
where $s$ is the scaling factor, $clamp$ function limits the value into the range of k bit integer $[-2^{k-1},2^{k-1}-1]$.
For 8 bit integer, the range is [-128,127].
$x_q$ can be de-quantized as $\hat{x}=sx_q\approx x$.

\textbf{Optimize Quantization Parameters}: 
The final step searches for the best quantization parameters to minimize the quantization error. 
This error is typically evaluated using a metric, such as MSE~\cite{choukroun2019lowbit_for_efficient_inference}, cosine distance~\cite{wu2020easyquant}, KL distance~\cite{migacz2017tensorrt}.
The optimization can be layer-wise~\cite{migacz2017tensorrt,yuan2022ptq4vit}, block-wise~\cite{nagel2020adaround,wei2022qdrop}, or network-wise~\cite{wang2022leveraging}.
For example, we can layer-wisely optimize the scaling factors $s^l$ in Eq~\ref{quantization_function} by minimize the MSE of de-quantized activation $\hat{X^l}$ and original activation $X^l$:
\begin{equation}
    \arg\min_{s^l_x}MSE(\hat{X^l},X^l).
\end{equation}
We can use various methods, such as grid search or gradient-based methods to solve the optimization problem.
The grid search method is a commonly used approach, which tests a range of candidate values for the quantization parameters and selects the parameters that minimize the quantization error.


\subsection{PTQ Reliability Evaluation Method}
\label{sec_evaluation_method}

The reliability of deep models is often measured through different dimensions, some commonly used dimensions include (1) model performance in various situations (including performance on the existing worst categories, noise samples, and out-of-distribution samples, etc.), (2) model robustness against test-time attacks such as adversarial attacks, and (3) the quality of the model's confidence. In this paper, we mainly focus on evaluating the reliability of existing PTQ methods based on the first dimension. Specifically, we measure its reliability by observing the accuracy changes of different categories under different quantization settings. Exploration of other dimensions will be addressed in future work.


To this end, we assess the prediction accuracy on different categories\footnote{Different categories can refer not only to different class labels in classification tasks but also to grouping of object detection bounding boxes based on their sizes, shapes, or other properties, as well as to grouping of samples based on their difficulty level. } to evaluate the reliability of PTQ. 
By analyzing the prediction accuracy across multiple categories, we can gain a more comprehensive understanding of the PTQ reliability and its ability to generalize across different types of inputs. 
This approach can provide additional insights into the performance of the quantized network and help identify any potential biases or limitations that may affect its overall reliability. 


To demonstrate the necessity of assessing on different categories, we experiment three representative networks, ResNet~\cite{he2016resnet} for CIFAR-10~\cite{CIFAR2009} classification task, MobileNetV2~\cite{sandler2018mobilenetv2} for ImageNet~\cite{Deng_ImageNet_CVPR2009} classification task, and YOLOv5~\cite{jocher2022yolov5} for MS COCO~\cite{lin2014mscoco} object detection task.
We performed multiple trials of network quantization with different random seeds.
In each trial, we randomly sampled a calibration dataset from the training dataset and performed PTQ quantization. 
We evaluated the performance of the quantized neural network in terms of both the average accuracy and the accuracy of each category.


\begin{figure}[tbp] 
    \centering
    \includegraphics[width=0.45\textwidth]{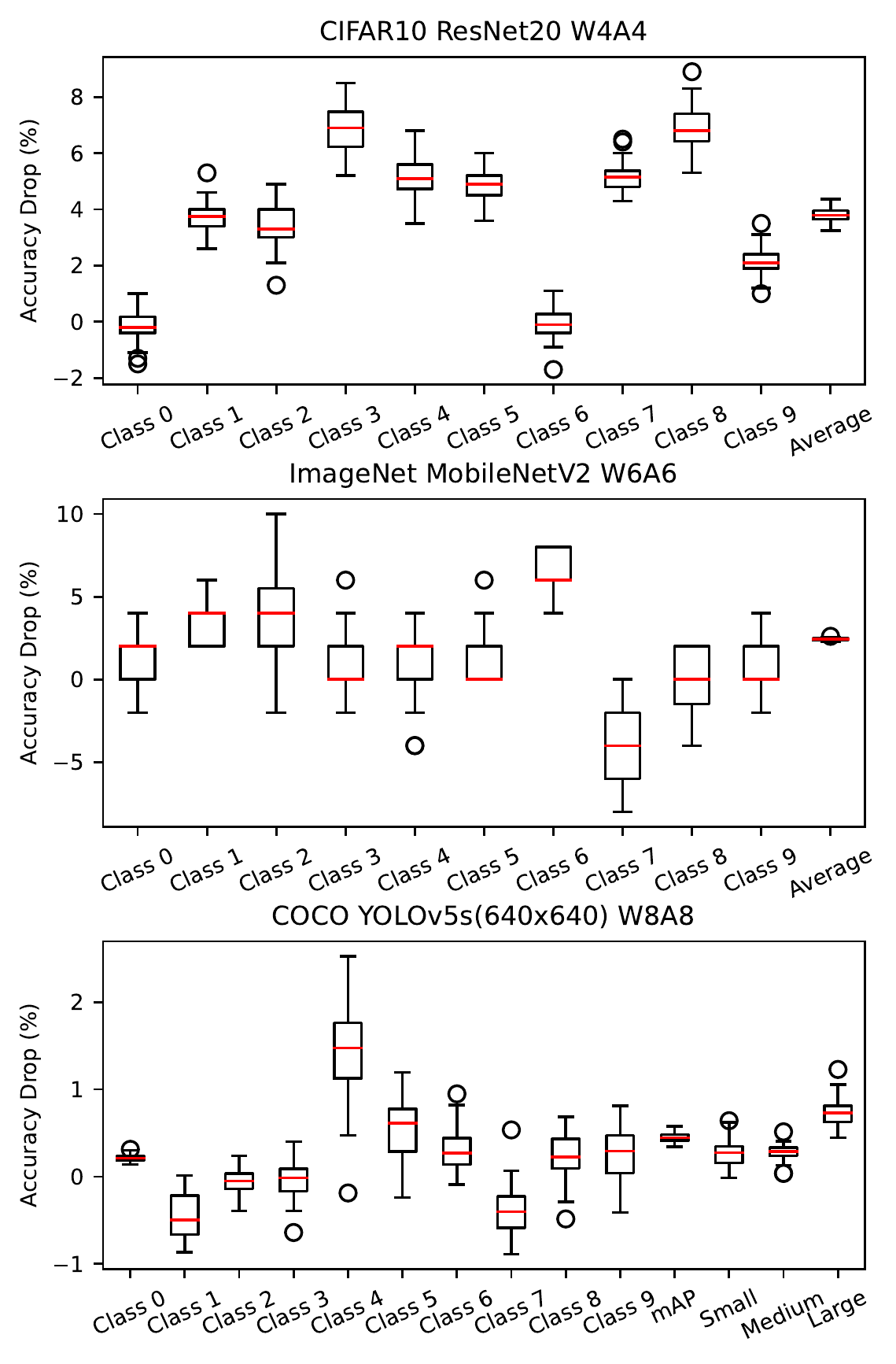} 
    \caption{The box plot of accuracy drop on different classes over 50 trials with different random seeds. We only plot 10 classes of ImageNet and COCO for demonstration. "mAP" refers to mean average precision. "Small", "Medium", and "Large" refer to the precision for small, medium, and large objects.}
    \label{im_data_intra_var}
\end{figure}

Figure~\ref{im_data_intra_var} illustrates the accuracy drop from the original network, with the box spanning from the first quartile (Q1) to the third quartile (Q3) of the data, while a red horizontal line depicts the median.
Notably, the accuracy drop exhibits significant difference across categories and average values.
Firstly, we observe that different categories display varying degrees of sensitivity to quantization. 
While some categories maintain their prediction accuracy after quantization, others experience a substantial decline in accuracy.
For instance, the accuracy drop of large objects is much larger than small and medium objects, indicating large objects are more vulnerable to quantization.
Secondly, we find that the variation in accuracy drop differs significantly across categories. 
For instance, the variation in Class 4 of COCO is much larger than that of other classes. 
Thirdly, we observe that the variance of accuracy drop across most individual categories is substantially higher than the average value. 
In conclusion, the reliability of individual categories is lower than anticipated.

Drawing on the above findings, we indicate that the low reliability of individual categories poses a risk in practical applications.
In many real-world scenarios, the accuracy of predictions for specific categories may be of critical importance, and any decrease in reliability for these categories can lead to serious consequences. 
For example, in medical diagnosis, the accuracy of predicting a disease may be more important than the overall accuracy of the model. 
Overall, this evaluation method can enhance the assessment of PTQ reliability and facilitate the deployment of PTQ in various real-world applications.

\section{Factors Affect PTQ Reliability}

In the preceding section, we propose an approach to assess the reliability of PTQ. 
Subsequently, we aim to examine how different factors involved in the quantization workflow impact the reliability of PTQ. 
To achieve this goal, we will conduct a series of tests on each step of PTQ workflow, including calibration dataset construction, quantization settings, and optimization of quantization parameters.


\subsection{Construction of Calibration Dataset}
\label{sec_calibration_dataset}


The calibration dataset is used to estimate the distribution of the activations in the network we want to quantize. 
If the calibration dataset fails to capture the statistical characteristics of the real-world data, the accuracy of the quantized network may decrease, since the quantization parameters are derived based on the distribution of the activations.
In this subsection, we analyze the impact of constructing a calibration dataset on the reliability of PTQ. 
We consider three factors that can affect the distribution of the calibration dataset: noise data, inter-class bias and dataset size.

\begin{figure}[tbp] 
    \centering
    \includegraphics[width=0.5\textwidth]{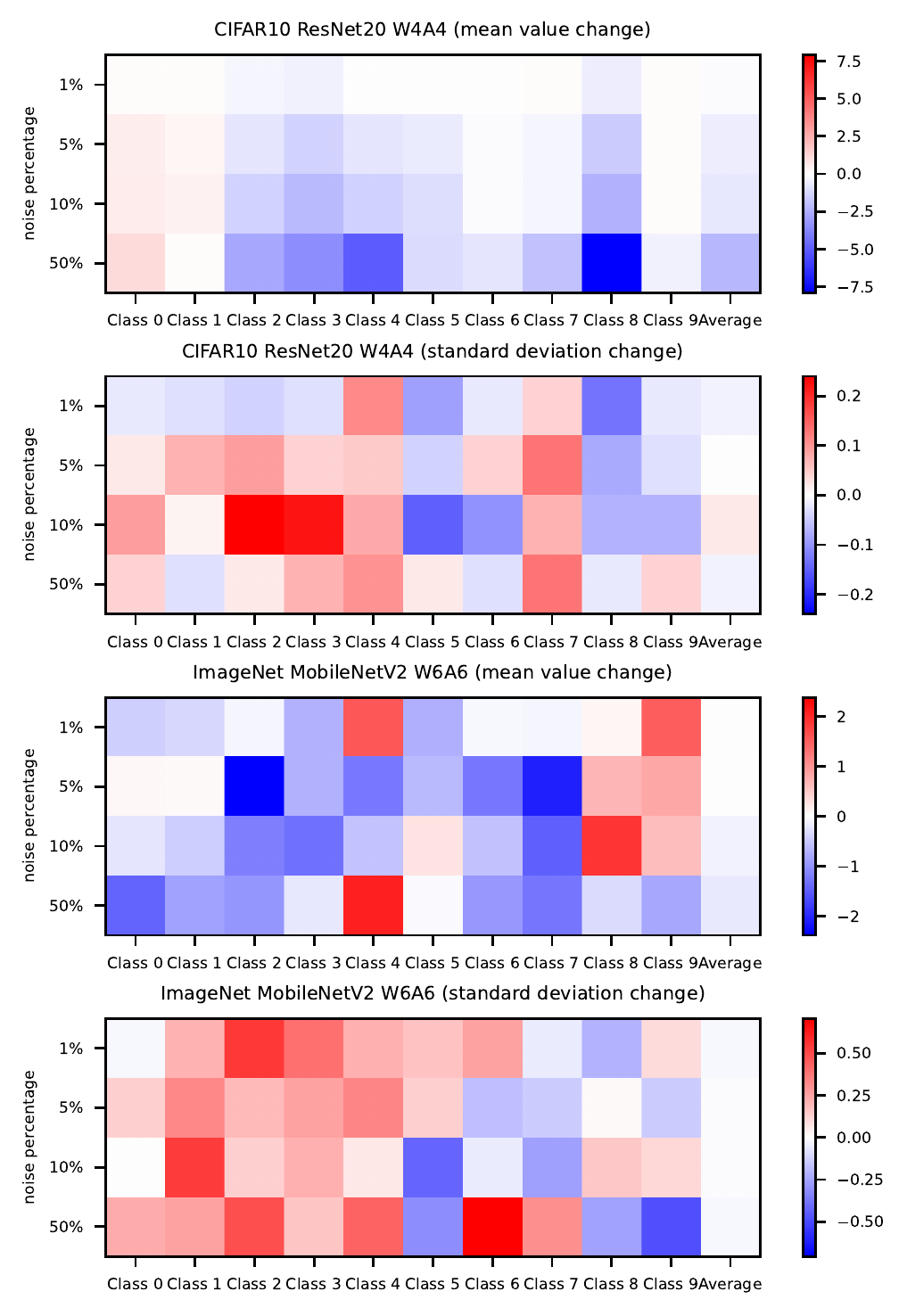} 
    \caption{Influence of noisy calibration data. This figure plots the relative performance change to the clean case with varying data noise amounts. The change values are demonstrated in different colors (red means accuracy increment while blue means decrement). We execute 50 trails for each percentage of noise.}
    \label{im_noise_data}
\end{figure}

\textbf{Noise data} refers to data samples that are not representative of the underlying distribution of the data. 
Including noise data in a calibration dataset can have a negative impact on PTQ, as the noise data can bias the distribution of activation and lead to inaccurate quantization parameters. 
To assess the impact of noise data, we construct the calibration dataset with some images sampled from training set and some randomly generated noise images.

Figure~\ref{im_noise_data} demonstrates the prediction accuracy mean value change and standard deviation change comparing with experiments without noise on the prediction accuracy.
We observe that increasing the percentage of noise data leads to a reduction on the average accuracy.
The more noise data, the more average accuracy drop.
However, the impact on individual classes is much larger than average accuracy. 
We observe that most of the classes are vulnerable to the noise data.
For instance, the prediction accuracy of Class 8 of CIFAR-10 drops significantly when noise data is introduced. 
While some classes, such as Class 1 and Class 9 of CIFAR-10, exhibit consistent predictive accuracy. 
Additionally, it is noteworthy that the noise data changes the standard deviation of individual classes, while the standard deviation on Average not changes too much.


\textbf{Inter-class bias} refers to the presence of differences in the distribution between classes. 
To evaluate the impact of inter-class bias on the quantization process, we construct unbalanced calibration datasets, where the number of samples from each class is different.
We build the unbalanced dataset by increasing the sampling probability of a specific class. 
Specifically, we constructed calibration datasets in which the sampling probability of a certain class was set to 50\%, while the remaining classes were included with equal probability.

\begin{figure}[tbp] 
    \centering
    \includegraphics[width=0.5\textwidth]{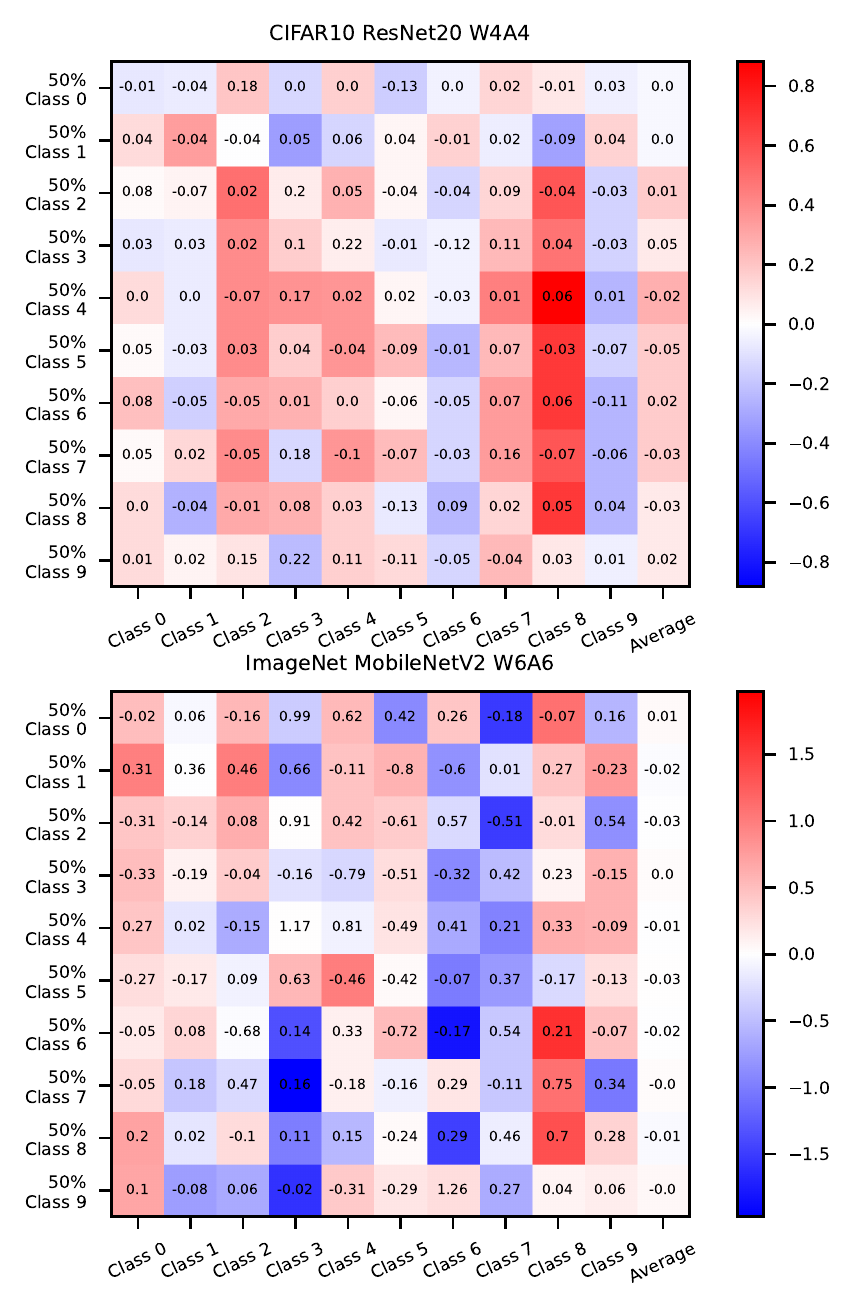} 
    \caption{Influence of unbalanceed calibration datasets. The reported top-1 accuracy is averaged over 50 runs with different random seeds. The prediction accuracy change is demonstrated in different colors (red means increment and blue means decrement). The standard deviation change is annotated as text (positive means increment and negative means decrement).}
    \label{im_data_cls_bias}
\end{figure}


Figure~\ref{im_data_cls_bias} depicts the results obtained using unbalanced calibration datasets. 
The figure demonstrates that increasing the number of different classes leads to slight change on the average prediction accuracy, while significant change on the prediction accuracy of individual classes.
The results also indicate that some classes are more susceptible to the unbalanced calibration dataset. 
For instance, the prediction accuracy of Class 6 on ImageNet significantly decreases. 
It is also worth noting that increasing the number of a certain class does not necessarily improve the prediction accuracy on this class. 
In addition, we note that the standard deviation of the average prediction accuracy almost remains unchanged, whereas the standard deviation of individual classes displays a marked variation.
The distribution of various classes may vary, and an unbalanced dataset can alter the distribution of the calibration dataset. 
Inter-class bias can impact the selection of quantization parameters, leading to variations in prediction accuracy. 
However, the average prediction accuracy remains relatively stable, indicating its robustness to changes in quantization parameters. 
Conversely, significant variations in the prediction accuracy of individual classes suggest their vulnerability to perturbations caused by changes in the calibration dataset.

\begin{table}[tb]
\centering
\caption{The influence of different numbers of calibration samples. We report the mean±std over 50 runs.}
\label{table_dataset_size}
\resizebox{\linewidth}{!}{
\begin{tabular}{@{}ccccc@{}}
\toprule
Dataset Size & 1         & 32        & 256       & 1024      \\ \midrule
\multicolumn{5}{c}{CIFAR-10 ResNet20 W4A4}                   \\ \midrule
Average      & 88.0±0.53 & 88.0±0.30 & 88.1±0.18 & 88.0±0.17 \\
Class 0      & 92.6±0.81 & 92.3±0.41 & 92.1±0.52 & 92.0±0.45 \\
Class 1      & 92.3±0.80 & 93.0±0.52 & 93.1±0.33 & 93.1±0.36 \\
Class 2      & 83.2±1.09 & 84.4±0.75 & 84.8±0.57 & 84.8±0.48 \\
Class 3      & 77.8±1.70 & 76.6±0.87 & 76.3±0.69 & 76.2±0.77 \\
Class 4      & 86.5±1.22 & 86.7±0.58 & 86.8±0.47 & 86.9±0.50  \\
Class 5      & 81.7±1.65 & 81.5±0.71 & 81.6±0.50 & 81.5±0.61 \\
Class 6      & 94.1±0.50 & 94.2±0.42 & 94.1±0.45 & 94.0±0.38 \\
Class 7      & 89.4±0.69 & 89.4±0.57 & 89.7±0.53 & 89.6±0.47 \\
Class 8      & 88.9±1.23 & 88.9±0.99 & 88.7±0.58 & 88.6±0.46 \\
Class 9      & 93.5±0.50 & 93.5±0.47 & 93.3±0.41 & 93.3±0.44 \\ \midrule
\multicolumn{5}{c}{Imagenet MobilenetV2 W6A6}                \\ \midrule
Average      & 70.0±0.38 & 70.2±0.06 & 70.2±0.05 & 70.2±0.07 \\
Class 0      & 82.8±1.44 & 83.1±1.34 & 83.3±1.48 & 82.8±1.13 \\
Class 1      & 68.8±2.22 & 70.4±1.34 & 70.2±1.40 & 70.4±1.15 \\
Class 2      & 72.8±2.71 & 74.4±2.58 & 74.6±2.31 & 73.8±2.77 \\
Class 3      & 74.6±2.33 & 73.6±2.04 & 74.8±1.83 & 74.3±2.28 \\
Class 4      & 61.2±2.11 & 60.8±1.92 & 61.2±1.65 & 60.8±1.79 \\
Class 5      & 95.1±1.28 & 95.2±1.12 & 95.0±1.15 & 95.2±1.26 \\
Class 6      & 84.4±2.53 & 85.3±1.90 & 85.4±1.79 & 85.2±1.21 \\
Class 7      & 63.5±2.49 & 64.6±2.06 & 65.9±2.12 & 65.4±2.37 \\
Class 8      & 80.8±1.54 & 80.0±1.52 & 80.1±1.92 & 80.0±1.62 \\
Class 9      & 91.8±1.90 & 91.7±1.74 & 91.3±1.68 & 91.6±1.80  \\ \bottomrule
\end{tabular}
}
\end{table}

\textbf{The size of calibration dataset} is an important factor for PTQ. 
It is generally understood that small dataset can lead to inaccurate quantization parameters, while a large dataset can increase the accuracy of the quantization, resulting in a stable and reliable result.
To evaluate the number of calibration samples, experiments with different numbers of samples are be conducted.
As shown in Table~\ref{table_dataset_size}, we observe that larger dataset size results in higher and more stable average prediction accuracy, aligning with the generally accepted understanding. 
As the calibration dataset size increases from 1 to 32, there is a significant reduction in the variance of prediction accuracy.
For instance, when size increases from 1 to 32, the standard deviation on Class 4 of CIFAR-10 decreases from 1.22 to 0.58.

However, we observe that increasing the dataset size beyond 32 has little effect on reducing the variance of accuracy on individual classes.
Despite increasing the calibration dataset size to 1024, there may still be significant variance in the accuracy of some classes.
For instance, the Class 2 of ImageNet has a standard deviation of 2.77 on prediction accuracy.
Some classes may inherently have more variability than others, and increasing the dataset size may not necessarily reduce this variability. 
Therefore, simply increasing the dataset size may not be sufficient to reduce the variance.

\subsection{Quantization Settings}
\label{sec_quantization_settings}

Once the calibration dataset is collected, the next step is to select appropriate quantization settings. 
In this paper, we only examine the effects of bit-width configurations on uniform quantization.
Mixed-precision quantization and non-uniform quantization are expected to be the future directions of research.

\begin{table}[t]
\centering
\caption{The influence of quantization settings. We report mean±std over 50 runs with different random seeds. W6A6 means 6-bit weight and 6-bit activation.}
\label{table_quantization_settings}
\resizebox{\linewidth}{!}{
\begin{tabular}{@{}ccccc@{}}
\toprule
Task     & \multicolumn{2}{c}{CIFAR-10 ResNet20} & \multicolumn{2}{c}{ImageNet MobilenetV2} \\
bit-width & W6A6              & W4A4              & W8A8                & W6A6               \\ \midrule
Average  & 90.66±0.08        & 88.01±0.21        & 72.0±0.03           & 70.2±0.07          \\
Class 0  & 90.14±0.27        & 92.18±0.43        & 84.0±0.00           & 82.8±1.27          \\
Class 1  & 95.77±0.19        & 93.22±0.45        & 74.1±1.44           & 70.6±1.13          \\
Class 2  & 88.54±0.30        & 84.55±0.61        & 76.4±1.00           & 74.3±2.53          \\
Class 3  & 82.64±0.51        & 76.42±0.69        & 77.4±0.90           & 74.7±1.82          \\
Class 4  & 90.76±0.27        & 86.56±0.61        & 62.4±1.44           & 60.6±1.89          \\
Class 5  & 85.51±0.33        & 81.40±0.67        & 95.9±0.39           & 95.2±1.38          \\
Class 6  & 92.14±0.28        & 94.19±0.45        & 94.0±0.00           & 85.6±1.56          \\
Class 7  & 93.19±0.29        & 89.52±0.46        & 62.0±0.28           & 65.7±2.09          \\
Class 8  & 93.15±0.21        & 88.84±0.70        & 82.0±0.00           & 79.9±1.60          \\
Class 9  & 94.82±0.21        & 93.23±0.43        & 93.6±0.77           & 91.4±1.56          \\ \bottomrule
\end{tabular}
}
\end{table}

We conducted experiments by setting the same bit-width for all layers in the network and tested the prediction accuracy across 50 trials.
The results are demonstrated in Table~\ref{table_quantization_settings}.
We observe that the bit-width have a significant impact on the performance of quantized networks. 
Specifically, a higher bit-width results in not only more accurate but also more stable quantized networks. 
For example, the mean and std of average prediction accuracy is 36.6±0.06 on W8A8, while that of W6A6 is 31.4±0.41 for YOLOv5s quantization. 

We also observe that the performance of individual classes varies from each other.
Some classes experience a significant drop in prediction accuracy, while others do not. 
For example, Class 6 of ImageNet experience a decrease of more than 6\% in prediction accuracy from W8A8 to W6A6, while Class 5 of ImageNet experience a decrease of less than 1\%. 
We think this is because different classes require different dynamic ranges to achieve a certain prediction accuracy. 
The reduced number of bits used for representing the network's parameters leads to a loss of information and reduced dynamic range.
When the bit-width drops, the dynamic range drops significantly. 
Consequently, the accuracy of some classes may decrease significantly.

It is worth noting that even a slight decrease in average prediction accuracy can result in substantial drops in both their accuracy and stability on certain categories. 
In order to maintain prediction accuracy and stability of quantization, it is important to exercise caution when decreasing the bit-width.

\subsection{Optimization of Quantization Parameters }
\label{sec_optimization}

The final step of PTQ is to search for the optimal quantization parameters.
The goal of this step is to minimize the quantization error, which is typically evaluated using a calibration metric, such as mean squared error (MSE)~\cite{choukroun2019lowbit_for_efficient_inference}, cosine distance~\cite{wu2020easyquant}, Kullback-Leibler (KL) divergence~\cite{migacz2017tensorrt}, or MinMax~\cite{nagel2021white}.

\begin{table}[tb]
\centering
\caption{The influence of calibration metric. We report mean±std over 50 runs with different random seeds.}
\label{table_optimization}
\resizebox{\linewidth}{!}{
\begin{tabular}{ccccc}
\hline
Metric  & MSE       & Cosine    & KL        & MinMax    \\ \hline
\multicolumn{5}{c}{CIFAR-10 ResNet20 W4A4}              \\ \hline
Average & 88.1±0.19 & 88.3±0.17 & 87.0±0.29 & 79.6±1.06 \\
Class 0 & 92.3±0.37 & 92.3±0.42 & 92.6±0.49 & 88.7±1.64 \\
Class 1 & 93.1±0.42 & 91.9±0.50 & 94.0±0.40 & 88.4±1.08 \\
Class 2 & 84.6±0.46 & 83.5±0.59 & 83.8±0.68 & 71.1±1.89 \\
Class 3 & 76.5±0.71 & 79.2±0.65 & 73.9±0.80 & 73.6±1.24 \\
Class 4 & 86.7±0.64 & 86.8±0.42 & 84.4±0.81 & 71.1±2.67 \\
Class 5 & 81.6±0.62 & 81.8±0.55 & 81.0±0.82 & 72.2±1.80 \\
Class 6 & 94.2±0.40 & 94.3±0.37 & 93.7±0.43 & 89.4±1.05 \\
Class 7 & 89.8±0.49 & 90.1±0.47 & 88.4±0.61 & 86.3±0.85 \\
Class 8 & 88.8±0.74 & 89.6±0.42 & 86.1±0.87 & 64.9±3.05 \\
Class 9 & 93.4±0.45 & 93.5±0.34 & 92.5±0.56 & 90.0±0.65 \\ \hline
\multicolumn{5}{c}{ImageNet MobilenetV2 W6A6}           \\ \hline
Average & 70.2±0.07 & 70.1±0.07 & 69.9±0.06 & 69.7±0.07 \\
Class 0 & 82.8±1.27 & 82.1±0.69 & 82.8±1.21 & 83.7±1.62 \\
Class 1 & 70.6±1.13 & 69.1±1.61 & 70.2±1.38 & 70.4±1.56 \\
Class 2 & 74.3±2.53 & 71.3±2.65 & 76.1±2.56 & 74.3±3.43 \\
Class 3 & 74.7±1.82 & 74.2±2.26 & 75.6±2.91 & 75.1±2.47 \\
Class 4 & 60.6±1.89 & 61.0±1.80 & 60.2±1.46 & 61.4±2.31 \\
Class 5 & 95.2±1.38 & 95.4±1.00 & 94.7±1.18 & 93.7±1.96 \\
Class 6 & 85.6±1.56 & 82.7±2.02 & 85.1±1.80 & 83.5±2.25 \\
Class 7 & 65.7±2.09 & 62.8±1.96 & 64.7±2.18 & 64.0±1.94 \\
Class 8 & 79.9±1.60 & 79.4±1.15 & 81.1±1.56 & 81.8±2.09 \\
Class 9 & 91.4±1.56 & 92.0±1.52 & 91.4±1.67 & 91.1±1.75 \\ \hline
\end{tabular}
}
\end{table}

We conduct experiments to assess the reliability of PTQ using different calibration metrics.
Table~\ref{table_optimization} shows the results. 
For CIFAR-10, the mean prediction accuracy achieved by KL and MinMax algorithms is inferior to that of MSE and Cosine, with correspondingly higher standard deviation. 
However, in the case of average prediction accuracy of ImageNet, all four metrics display comparable accuracy and standard deviation.
For individual class, variations in performance can be observed across different metrics. 
To illustrate, when considering ImageNet, the MinMax metric exhibits higher prediction accuracy for Class 0, while MSE outperforms the other metrics for Class 1, and KL yields better results for Class 2, and Cosine performs better for Class 9. 
A high standard deviation on individual classes is observed across all four metrics. 
Additionally, it is noted that the variances obtained using the KL and MinMax metrics are greater than those obtained using other metrics.



The observed differences in performance among different calibration metrics could be attributed to the characteristics of each metric. 
One possible reason for the larger variance of the KL could be the existence of long tailed samples. 
KL divergence is a measure of the difference between two probability distributions, and is particularly sensitive to differences in tail behavior. 
This means that even small differences in the tails of the distributions can result in a large divergence value. 
The MinMax metric determines the quantization parameters based on the maximum value of the data in the calibration activation.
As a result, the MinMax metric may be more sensitive to outliers or extreme values in the calibration dataset, leading to a larger variance in the quantization error. 
In contrast, other metrics such as MSE and cosine distance may be more robust to outliers and therefore have a smaller variance. 
Further research is needed to fully understand the reasons behind these observations.



Another factor that can affect the reliability is the optimization algorithm.
In our study, we assessed four different optimization algorithms, including the grid search method~\cite{nagel2021white}, as well as three gradient-based approaches\footnote{These gradient based methods will optimize the scaling factors for quantizing activation and the rounding for quantizing weight.}, namely Adaround~\cite{nagel2020adaround}, BRECQ~\cite{li2021brecq}, and QDrop~\cite{wei2022qdrop}.
The results, as shown in Table~\ref{table_optimization_algorithm}, demonstrate that gradient-based methods can substantially enhance the overall prediction accuracy. 
For instance, the average accuracy on ImageNet using QDrop is 72.1\%, which is almost comparable to full-precision. 
However, our analysis reveals that the accuracy of individual classes varied considerably. 
For instance, using QDrop, Class 1 of ImageNet achieves a mean accuracy of only 68.6\%, whereas this is 70.6\% using the grid search algorithm. 
Additionally, our observation reveals that the standard deviations of individual classes are relatively high.
Therefore, we indicate that gradient-based PTQ methods may have relatively lower reliability, despite achieving better overall prediction accuracy.


\begin{table}[tb]
\centering
\caption{The influence of optimization algorithm. We report mean±std over 50 runs with different random seeds.}
\label{table_optimization_algorithm}
\resizebox{\linewidth}{!}{
\begin{tabular}{@{}ccccc@{}}
\toprule
Algorithm & Grid Search & Adaround  & BRECQ     & QDrop     \\ \midrule
\multicolumn{5}{c}{CIFAR10 ResNet20 W4A4}                   \\ \midrule
Average   & 88.0±0.21   & 89.8±0.14 & 90.0±0.14 & 89.3±0.22 \\
Class 0   & 92.2±0.43   & 93.1±0.41 & 91.4±0.54 & 90.1±0.88 \\
Class 1   & 93.2±0.45   & 95.5±0.28 & 95.3±0.41 & 96.3±0.47 \\
Class 2   & 84.6±0.61   & 87.9±0.50 & 87.2±0.61 & 88.0±0.79 \\
Class 3   & 76.4±0.69   & 82.0±0.76 & 78.8±0.78 & 79.6±1.21 \\
Class 4   & 86.6±0.61   & 90.1±0.62 & 89.7±0.50 & 88.0±0.88 \\
Class 5   & 81.4±0.67   & 81.8±0.78 & 83.4±0.65 & 81.0±1.13 \\
Class 6   & 94.2±0.45   & 91.3±0.45 & 92.9±0.48 & 90.5±0.81 \\
Class 7   & 89.5±0.46   & 92.0±0.46 & 92.9±0.38 & 93.1±0.53 \\
Class 8   & 88.8±0.70   & 92.5±0.41 & 93.6±0.43 & 92.6±0.57 \\
Class 9   & 93.2±0.43   & 93.4±0.42 & 94.5±0.34 & 93.4±0.61 \\ \midrule
\multicolumn{5}{c}{ImageNet MobileNetV2 W6A6}               \\ \midrule
Average   & 70.2±0.07   & 72.0±0.06 & 72.1±0.06 & 72.1±0.06 \\
Class 0   & 82.8±1.27   & 82.3±1.05 & 81.7±0.98 & 82.1±1.01 \\
Class 1   & 70.6±1.13   & 67.2±2.77 & 69.2±2.27 & 68.8±2.33 \\
Class 2   & 74.3±2.53   & 81.9±2.51 & 79.7±2.37 & 80.6±2.95 \\
Class 3   & 74.7±1.82   & 76.4±2.52 & 77.0±2.01 & 78.0±2.73 \\
Class 4   & 60.6±1.89   & 63.8±2.56 & 61.1±2.34 & 60.6±2.51 \\
Class 5   & 95.2±1.38   & 95.2±0.99 & 95.4±0.98 & 95.4±1.09 \\
Class 6   & 85.6±1.56   & 92.3±1.61 & 90.7±1.43 & 89.7±2.26 \\
Class 7   & 65.7±2.09   & 64.0±1.95 & 62.8±1.65 & 63.0±1.61 \\
Class 8   & 79.9±1.60   & 80.4±1.33 & 80.8±1.38 & 78.4±0.80 \\
Class 9   & 91.4±1.56   & 93.3±1.42 & 93.5±1.25 & 91.0±1.66 \\ \bottomrule
\end{tabular}
}
\end{table}
\section{Benchmark of PTQ Reliability}

We have presented the process for measuring the reliability of PTQ and systematically evaluated the impact of each step in the PTQ workflow on reliability. 
We have found that PTQ performance varies greatly across different networks and tasks, and it is not possible for us to present and analyze each of them in this paper. 
In addition, there are still many aspects of PTQ reliability that require further analysis and exploration. 
Therefore, we have developed a framework that can test the PTQ reliability of different networks according to the proposed approach in this paper.
As shown in Figure~\ref{im_overview}, the framework provide a standardized approach for evaluating the reliability of PTQ by assessing both the average performance and performance on various cetegories.
We have used this framework to test the PTQ reliability performance on various tasks, PTQ methods, and network architectures\footnote{The results are provided in supplementary materials.}, which can serve as a benchmark.
Furthermore, the framework can be extended to evaluate the reliability of PTQ on other datasets or tasks, as well as to investigate the effect of different calibration metrics or other parameters on the quantization accuracy. 
We hope that the proposed framework and the benchmark can contribute to the development and improvement of PTQ and other quantization methods, and ultimately enable more efficient deployment of deep learning models on more applications.
\section{Conclusion}
In this paper, we have introduced the concept of reliability in the context of post-training quantization and have proposed an approach to assess the reliability of PTQ. 
We have explored the impact of various factors on the reliability of PTQ, including the three main steps of the PTQ workflow: calibration dataset construction, quantization settings assignment, and quantization parameter optimization.
Furthermore, we have proposed a standardized framework for evaluating the reliability of PTQ on different neural networks, which can aid future research in this area.
Nevertheless, further investigation is necessary to fully comprehend the reliability of PTQ.


{\small
\bibliographystyle{ieee_fullname}
\bibliography{egbib}
}
\newpage
\section*{Appendix}










\begin{appendices}

\section{Quantization Models}
We have quantized various CNN architectures for the ImageNet dataset, such as ResNet-18, ResNet-50, RegNetX-600m, RegNetX-3200m, MobileNetV2, and MNasNet. 
Additionally, for the CIFAR-10 dataset, we considered ResNet models with different depths, including ResNet-20, ResNet-32, ResNet-44, and ResNet-56.
In our benchmark, we conducted thorough experiments on all the mentioned models to determine that quantization reliability issues exist in various networks.

\section{Benchmark}
\subsection{Evaluation on accuracy drop}
We quantize each model 50 times, selecting different random calibration data each time, and investigate the change in quantized model's accuracy for each class compared to the accuracy of the full-precision model.
The box plot of accuracy drop on different classes over 50 trials with different random seeds is shown as follows. 
We only plot 10 classes of ImageNet.

\begin{figure}[htb]
    \centering
    \includegraphics[width=0.9\linewidth]{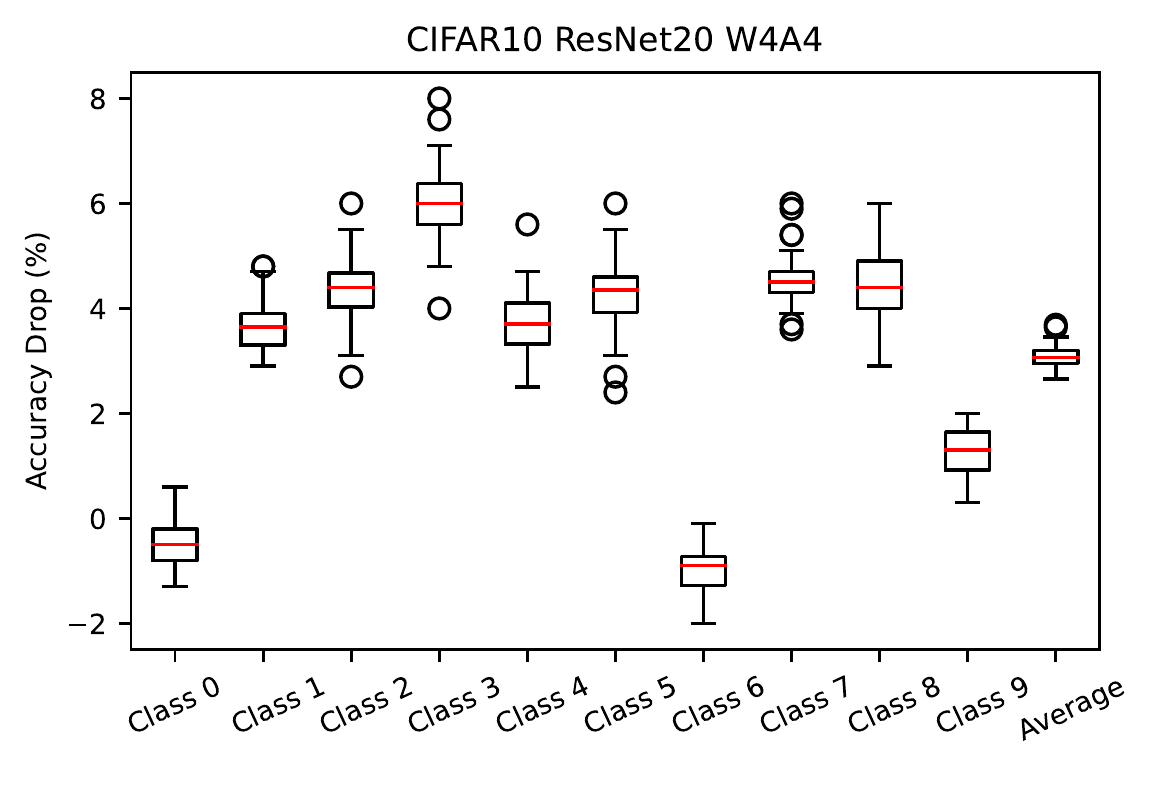} 
\end{figure}
\begin{figure}[htb]
    \centering
    \includegraphics[width=0.9\linewidth]{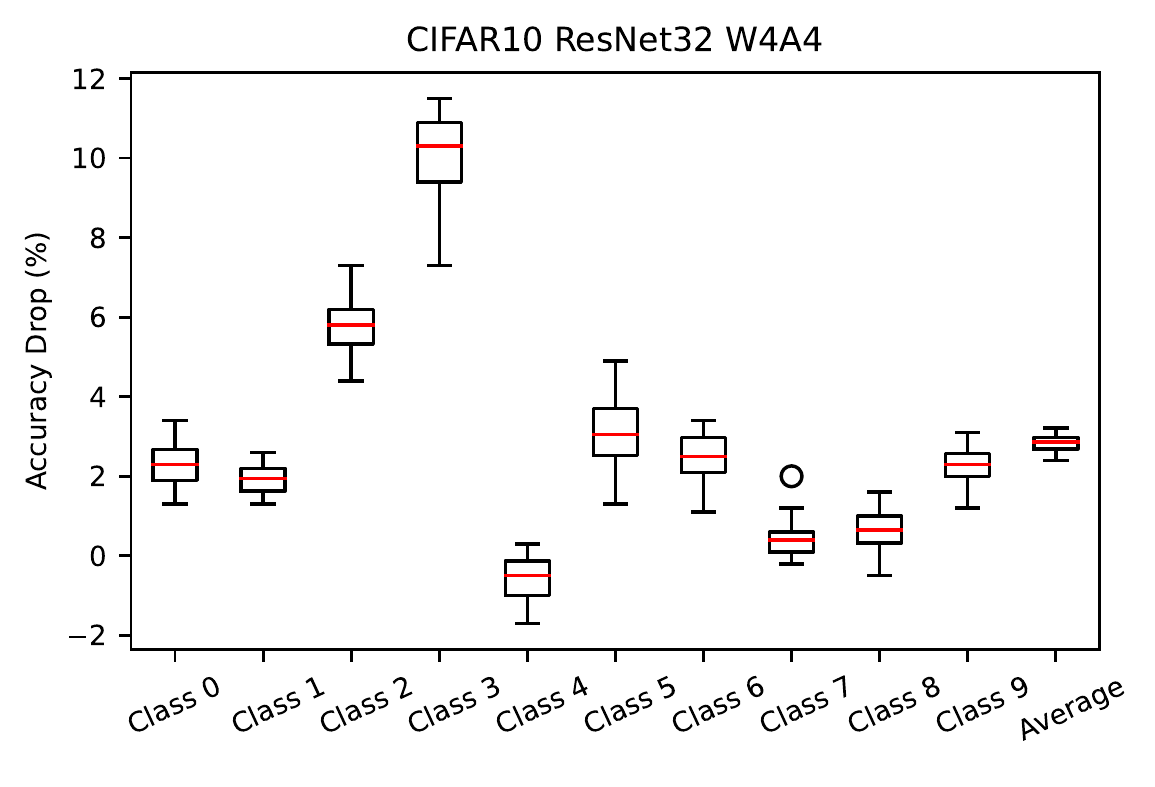}
\end{figure}

\begin{figure}[H]
    \centering
    \includegraphics[width=0.9\linewidth]{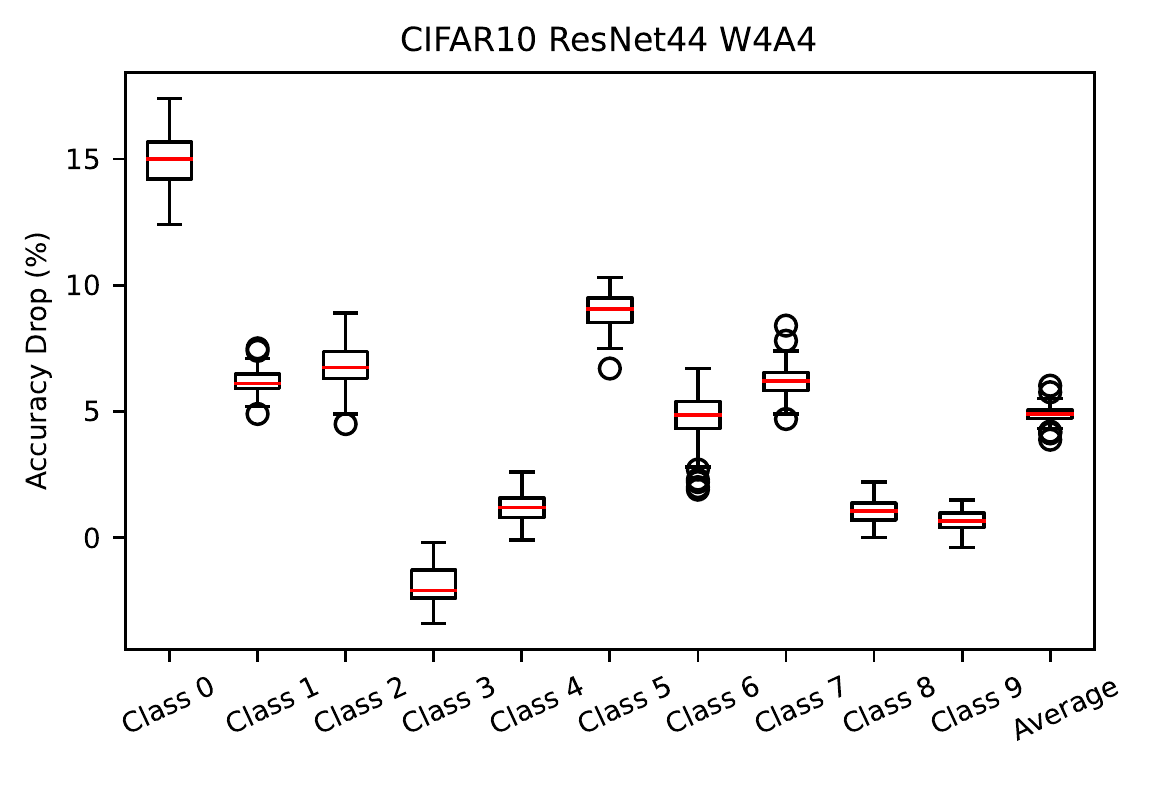} 
\end{figure}
\vspace{-1cm}
\begin{figure}[H]
    \centering
    \includegraphics[width=0.9\linewidth]{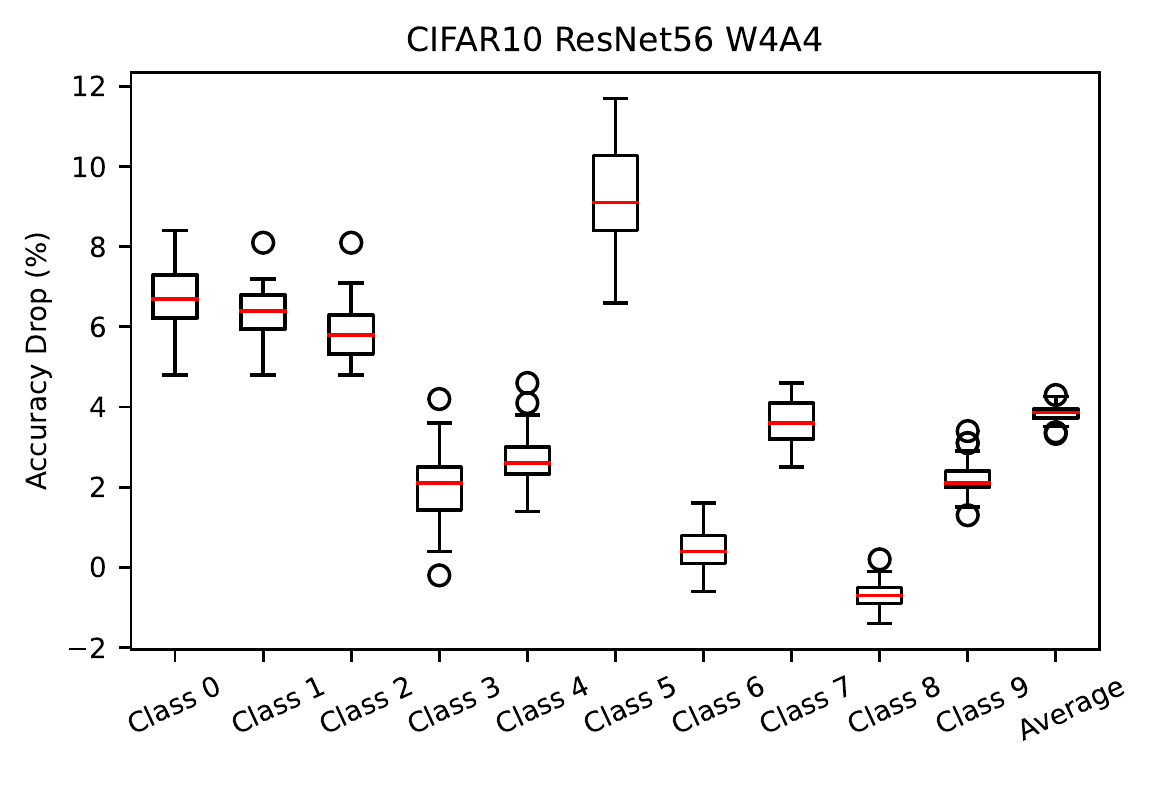} 
\end{figure}
\vspace{-1cm}
\begin{figure}[H]
    \centering
    \includegraphics[width=0.9\linewidth]{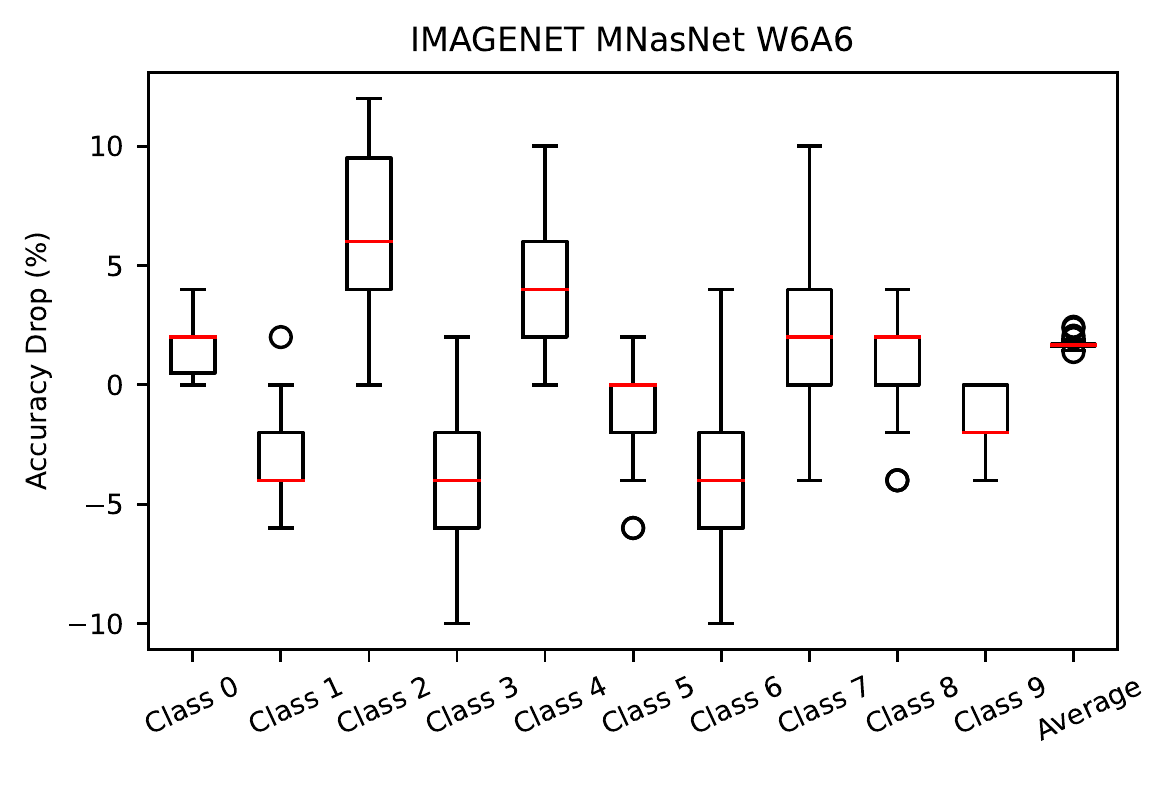} 
\end{figure}
\vspace{-1cm}
\begin{figure}[H]
    \centering
    \includegraphics[width=0.9\linewidth]{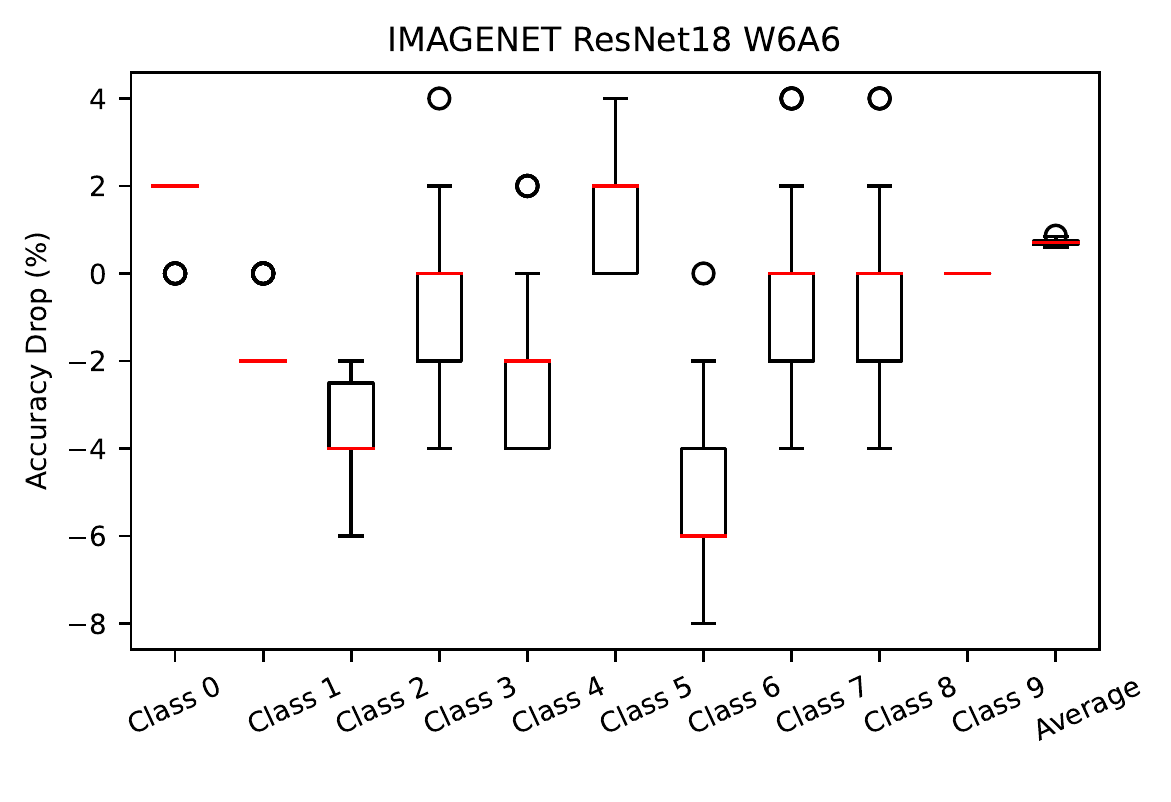}
\end{figure}

\begin{figure}[H]
    \centering
    \includegraphics[width=0.9\linewidth]{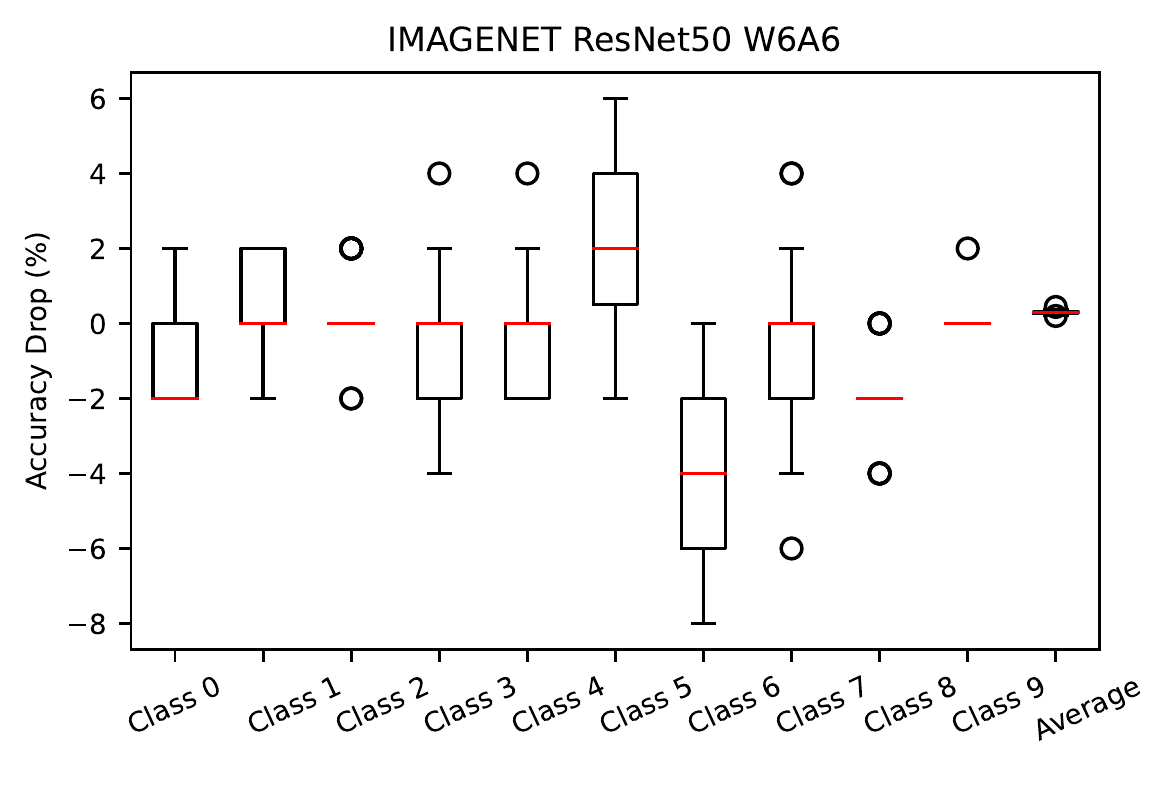}
\end{figure}
\vspace{-1cm}
\begin{figure}[H]
    \centering
    \includegraphics[width=0.9\linewidth]{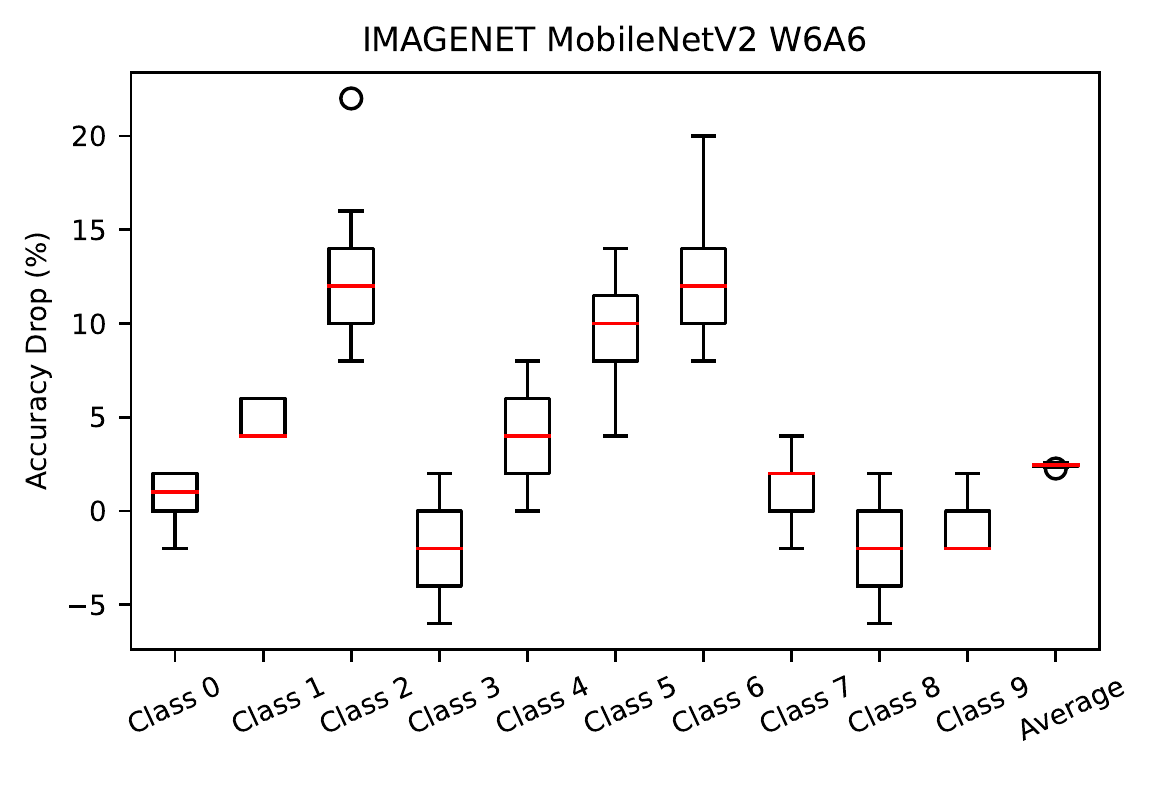}
\end{figure}
\vspace{-1cm}
\begin{figure}[H]
    \centering
    \includegraphics[width=0.9\linewidth]{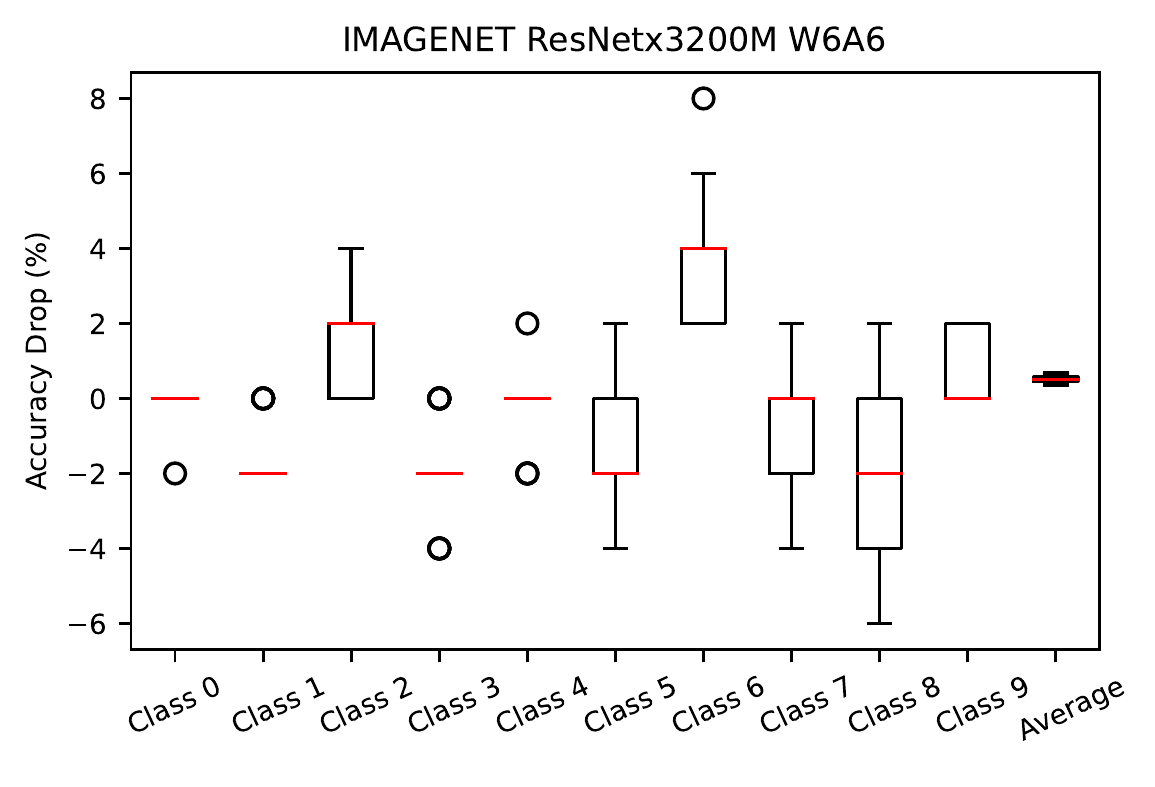}
\end{figure}
\vspace{-1cm}
\begin{figure}[H]
    \centering
    \includegraphics[width=0.9\linewidth]{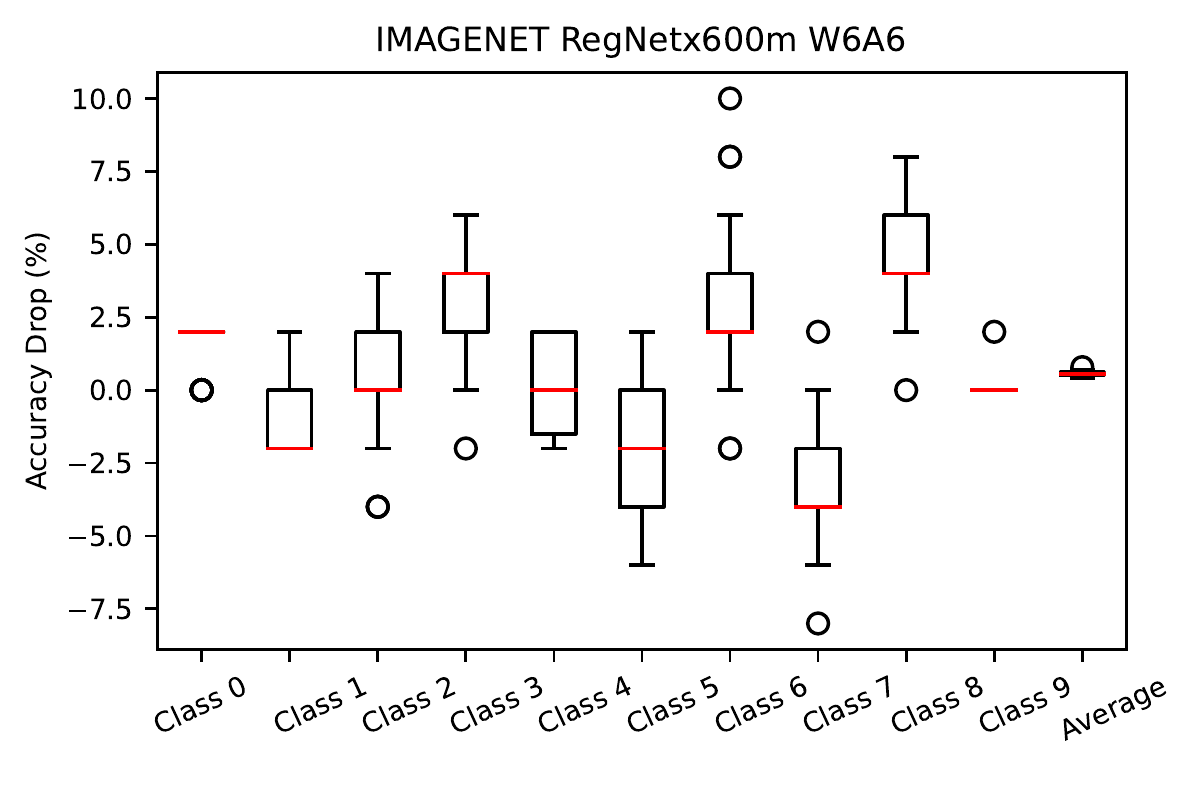}
\end{figure}

\subsection{Evaluation on calibration numbers}
We tested the performance of PTQ at calibration set sizes of 1, 32, 64, and 256 separately, and the experimental results of different models are as follows:

\begin{table}[tb]
\centering
\caption{The influence of different numbers of calibration samples. We report the mean±std over 50 runs on CIFAR-10 dataset.}
\resizebox{\linewidth}{!}{
\begin{tabular}{@{}ccccc@{}}
\toprule
Dataset Size & 1         & 32        & 256       & 1024      \\ \midrule
\multicolumn{5}{c}{CIFAR-10 ResNet20 W4A4}                   \\ \midrule
Average & 88.0±0.53 & 88.0±0.30 & 88.0±0.21 & 88.1±0.18 \\ 
Class 0 & 92.6±0.81 & 92.3±0.41 & 92.2±0.44 & 92.1±0.52 \\ 
Class 1 & 92.3±0.80 & 93.0±0.52 & 93.2±0.46 & 93.1±0.33 \\ 
Class 2 & 83.2±1.09 & 84.4±0.75 & 84.6±0.62 & 84.7±0.57 \\ 
Class 3 & 77.8±1.70 & 76.6±0.87 & 76.4±0.70 & 76.3±0.69 \\ 
Class 4 & 86.5±1.22 & 86.7±0.58 & 86.6±0.61 & 86.8±0.47 \\ 
Class 5 & 81.7±1.65 & 81.5±0.71 & 81.4±0.67 & 81.6±0.50 \\ 
Class 6 & 94.1±0.50 & 94.2±0.42 & 94.2±0.45 & 94.1±0.45 \\ 
Class 7 & 89.4±0.69 & 89.4±0.57 & 89.5±0.47 & 89.7±0.53 \\ 
Class 8 & 88.9±1.23 & 88.9±0.99 & 88.8±0.71 & 88.7±0.58 \\ 
Class 9 & 93.5±0.50 & 93.5±0.47 & 93.2±0.43 & 93.3±0.41 \\ \midrule
\multicolumn{5}{c}{CIFAR-10 ResNet32 W4A4}                \\ \midrule
        Average & 87.8±0.62 & 87.8±0.29 & 87.7±0.21 & 87.7±0.24 \\ 
        Class 0 & 91.5±0.88 & 91.7±0.64 & 91.8±0.54 & 91.7±0.55 \\ 
        Class 1 & 93.5±0.71 & 93.8±0.39 & 93.8±0.36 & 94.0±0.46 \\ 
        Class 2 & 84.9±1.24 & 86.0±0.52 & 86.1±0.63 & 86.0±0.58 \\ 
        Class 3 & 69.8±2.39 & 68.1±1.06 & 67.4±1.02 & 67.3±0.85 \\ 
        Class 4 & 90.5±1.08 & 90.4±0.63 & 90.3±0.49 & 90.3±0.67 \\ 
        Class 5 & 81.7±1.80 & 81.4±0.86 & 81.4±0.77 & 81.1±0.60 \\ 
        Class 6 & 91.9±0.91 & 92.2±0.55 & 92.3±0.54 & 92.4±0.58 \\ 
        Class 7 & 88.8±0.77 & 89.4±0.55 & 89.4±0.45 & 89.3±0.47 \\ 
        Class 8 & 93.5±0.90 & 93.2±0.60 & 93.2±0.52 & 93.1±0.44 \\ 
        Class 9 & 92.0±0.67 & 91.5±0.50 & 91.5±0.40 & 91.5±0.38 \\ \midrule
\multicolumn{5}{c}{CIFAR-10 ResNet44 W4A4}                \\ \midrule
        Average & 87.5±1.10 & 87.4±0.47 & 87.3±0.39 & 87.2±0.17 \\ 
        Class 0 & 77.5±3.07 & 77.6±1.63 & 77.4±1.13 & 77.1±0.79 \\ 
        Class 1 & 90.5±1.34 & 90.7±0.57 & 90.9±0.52 & 90.8±0.50 \\ 
        Class 2 & 84.1±2.14 & 84.0±1.00 & 83.9±0.97 & 83.6±0.68 \\ 
        Class 3 & 87.2±1.35 & 88.3±0.94 & 88.3±0.80 & 88.8±0.56 \\ 
        Class 4 & 91.7±0.88 & 91.2±0.61 & 91.2±0.55 & 91.3±0.51 \\ 
        Class 5 & 79.0±1.90 & 78.4±0.76 & 78.2±0.79 & 77.7±0.51 \\ 
        Class 6 & 89.4±2.36 & 88.0±1.55 & 87.7±1.16 & 87.2±0.57 \\ 
        Class 7 & 88.1±1.25 & 87.8±0.73 & 87.5±0.73 & 87.5±0.50 \\ 
        Class 8 & 94.0±1.16 & 94.1±0.59 & 94.2±0.52 & 94.2±0.38 \\ 
        Class 9 & 94.1±0.71 & 94.0±0.42 & 94.1±0.42 & 94.2±0.27 \\ \midrule
\multicolumn{5}{c}{CIFAR-10 ResNet56 W4A4}                \\ \midrule
        Average & 89.3±0.80 & 89.4±0.19 & 89.4±0.19 & 89.3±0.16 \\ 
        Class 0 & 86.6±2.03 & 86.8±1.00 & 86.8±0.82 & 86.7±1.00 \\ 
        Class 1 & 90.9±1.55 & 91.3±0.44 & 91.2±0.58 & 91.0±0.54 \\ 
        Class 2 & 84.2±1.53 & 85.2±0.78 & 85.2±0.68 & 85.5±0.58 \\ 
        Class 3 & 85.1±1.53 & 84.6±0.96 & 84.8±0.84 & 84.8±0.71 \\ 
        Class 4 & 91.5±0.96 & 91.6±0.72 & 91.6±0.61 & 91.6±0.67 \\ 
        Class 5 & 78.2±1.94 & 78.0±1.10 & 77.9±1.13 & 77.4±0.92 \\ 
        Class 6 & 94.8±0.63 & 94.6±0.44 & 94.7±0.52 & 94.7±0.36 \\ 
        Class 7 & 91.7±1.33 & 92.3±0.49 & 92.1±0.51 & 92.0±0.46 \\ 
        Class 8 & 96.3±0.49 & 96.4±0.31 & 96.4±0.31 & 96.5±0.33 \\ 
        Class 9 & 93.6±0.68 & 93.5±0.51 & 93.4±0.43 & 93.4±0.42 \\ \bottomrule
\end{tabular}
}
\end{table}

\begin{table}[htb]
\centering
\caption{The influence of different numbers of calibration samples. We report the mean±std over 50 runs on ImageNet dataset.}
\resizebox{\linewidth}{!}{
\begin{tabular}{@{}ccccc@{}}

\toprule
Dataset Size & 1         & 32        & 256       & 1024      \\ \midrule
\multicolumn{5}{c}{ImageNet ResNet18 W6A6}                   \\ \midrule
        Average & 70.0±0.29 & 70.3±0.07 & 70.3±0.06 & 70.3±0.05 \\ 
        Class 0 & 85.0±1.08 & 84.2±0.65 & 84.3±0.69 & 84.3±0.73 \\ 
        Class 1 & 89.2±1.39 & 89.6±0.83 & 89.5±0.85 & 89.4±0.93 \\ 
        Class 2 & 83.9±0.84 & 83.4±0.98 & 83.6±1.0 & 83.3±1.18 \\ 
        Class 3 & 68.6±2.02 & 68.4±1.46 & 68.6±1.85 & 68.4±1.71 \\ 
        Class 4 & 90.8±1.44 & 90.4±1.73 & 90.0±1.85 & 89.9±1.7 \\ 
        Class 5 & 70.0±1.52 & 70.8±1.26 & 70.9±1.14 & 70.7±1.11 \\ 
        Class 6 & 79.0±1.97 & 79.3±1.48 & 79.2±1.64 & 78.8±1.39 \\ 
        Class 7 & 69.4±1.39 & 68.5±1.9 & 68.5±2.26 & 68.9±1.97 \\ 
        Class 8 & 86.4±2.34 & 87.9±1.72 & 88.1±1.76 & 88.5±1.54 \\ 
        Class 9 & 98.5±1.17 & 99.9±0.39 & 100.0±0.0 & 100.0±0.0 \\ \midrule
\multicolumn{5}{c}{ImageNet ResNet50 W6A6}                \\ \midrule
        Average & 76.2±0.3 & 76.3±0.05 & 76.3±0.04 & 76.3±0.05 \\ 
        Class 0 & 94.2±0.86 & 94.7±1.04 & 95.0±1.08 & 95.0±1.15 \\ 
        Class 1 & 89.8±2.21 & 93.1±1.15 & 93.2±1.12 & 93.5±1.17 \\ 
        Class 2 & 81.7±1.22 & 81.6±1.06 & 81.7±0.9 & 81.7±0.9 \\ 
        Class 3 & 83.8±1.77 & 83.8±1.94 & 84.2±1.51 & 84.3±1.5 \\ 
        Class 4 & 87.0±1.08 & 86.5±1.24 & 86.3±1.44 & 86.0±1.41 \\ 
        Class 5 & 75.1±2.27 & 75.8±2.29 & 75.6±2.37 & 75.2±1.96 \\ 
        Class 6 & 81.5±1.91 & 81.8±1.53 & 82.0±1.83 & 81.5±1.73 \\ 
        Class 7 & 71.8±2.71 & 72.4±2.36 & 72.8±2.0 & 73.1±2.2 \\ 
        Class 8 & 84.3±1.06 & 83.6±1.08 & 84.0±0.94 & 83.2±1.32 \\ 
        Class 9 & 99.9±0.62 & 100.0±0.28 & 100.0±0.28 & 100.0±0.28 \\ \midrule
\multicolumn{5}{c}{ImageNet MobileNetV2 W6A6}                \\ \midrule
        Average & 70.0±0.37 & 70.2±0.06 & 70.2±0.06 & 70.2±0.05 \\ 
        Class 0 & 90.5±1.1 & 91.0±1.08 & 91.0±1.08 & 91.0±1.08 \\ 
        Class 1 & 83.2±2.91 & 85.2±1.13 & 85.3±0.95 & 85.3±1.03 \\ 
        Class 2 & 74.4±2.34 & 74.2±2.46 & 74.0±2.38 & 73.3±2.18 \\ 
        Class 3 & 78.2±1.78 & 78.5±1.82 & 78.0±1.83 & 78.1±1.76 \\ 
        Class 4 & 79.0±1.8 & 78.1±1.92 & 78.2±1.87 & 77.9±1.92 \\ 
        Class 5 & 67.9±2.53 & 68.3±2.15 & 68.4±2.01 & 67.9±1.7 \\ 
        Class 6 & 69.4±2.54 & 69.2±2.66 & 69.9±2.31 & 69.8±2.01 \\ 
        Class 7 & 73.4±1.27 & 73.5±1.42 & 73.1±1.39 & 73.2±1.33 \\ 
        Class 8 & 88.1±2.43 & 88.5±1.59 & 87.9±2.09 & 88.6±1.67 \\ 
        Class 9 & 97.8±0.65 & 97.3±1.03 & 97.3±1.03 & 97.3±1.24 \\ \midrule
\multicolumn{5}{c}{ImageNet MNasNet W6A6}                \\ \midrule
        Average & 74.9±0.2 & 74.8±0.22 & 74.8±0.19 & 74.8±0.14 \\ 
        Class 0 & 95.5±1.19 & 96.0±1.44 & 96.0±1.41 & 95.9±1.2 \\ 
        Class 1 & 91.0±1.56 & 91.0±1.66 & 91.2±1.59 & 91.0±1.66 \\ 
        Class 2 & 81.2±2.66 & 80.3±2.05 & 79.5±2.87 & 79.6±2.7 \\ 
        Class 3 & 83.1±2.86 & 82.6±2.77 & 82.4±2.95 & 82.1±2.5 \\ 
        Class 4 & 80.8±2.62 & 82.0±2.91 & 82.0±2.21 & 82.0±2.21 \\ 
        Class 5 & 79.2±2.23 & 78.2±2.58 & 79.1±2.01 & 79.2±2.65 \\ 
        Class 6 & 79.0±3.28 & 79.5±2.29 & 79.9±2.65 & 80.2±2.46 \\ 
        Class 7 & 65.4±3.5 & 66.8±3.73 & 67.7±3.5 & 69.2±2.53 \\ 
        Class 8 & 90.0±2.19 & 89.0±2.31 & 88.9±2.3 & 88.5±2.49 \\ 
        Class 9 & 97.9±1.41 & 97.9±1.09 & 97.6±1.13 & 97.5±1.1 \\ 
        \bottomrule
\end{tabular}
}
\end{table}

\begin{table}[htb]
\centering
\caption{The influence of different numbers of calibration samples. We report the mean±std over 50 runs on ImageNet dataset.}
\resizebox{\linewidth}{!}{
\begin{tabular}{@{}ccccc@{}}

\toprule
Dataset Size & 1         & 32        & 256       & 1024      \\ \midrule
\multicolumn{5}{c}{ImageNet RegNetx600M W6A6}                   \\ \midrule
        Average & 72.7±0.44 & 73.0±0.07 & 73.0±0.08 & 73.0±0.06 \\ 
        Class 0 & 90.4±1.4 & 90.7±1.04 & 90.4±0.77 & 90.3±0.93 \\ 
        Class 1 & 89.3±1.24 & 89.4±1.16 & 89.3±1.11 & 89.7±0.84 \\ 
        Class 2 & 75.2±1.49 & 75.9±2.09 & 75.8±1.86 & 76.6±1.57 \\ 
        Class 3 & 76.0±1.79 & 77.1±1.61 & 77.0±1.84 & 77.1±1.84 \\ 
        Class 4 & 90.0±1.79 & 89.1±1.84 & 89.9±1.49 & 89.9±1.85 \\ 
        Class 5 & 77.1±1.92 & 75.6±1.89 & 75.8±1.99 & 75.3±1.54 \\ 
        Class 6 & 76.2±2.55 & 76.9±2.72 & 77.0±2.37 & 77.5±2.55 \\ 
        Class 7 & 73.8±2.55 & 75.0±2.37 & 75.0±2.01 & 76.2±1.91 \\ 
        Class 8 & 83.0±2.44 & 83.8±1.64 & 83.4±1.79 & 83.3±1.86 \\ 
        Class 9 & 96.0±0.28 & 96.0±0.0 & 96.0±0.28 & 96.0±0.0 \\ \midrule
\multicolumn{5}{c}{ImageNet RegNetx3200M W6A6}                \\ \midrule
        Average & 77.8±0.21 & 77.9±0.07 & 77.9±0.07 & 77.9±0.06 \\ 
        Class 0 & 94.1±0.39 & 94.0±0.28 & 94.0±0.28 & 94.0±0.0 \\ 
        Class 1 & 90.9±1.07 & 91.8±0.6 & 91.8±0.65 & 91.9±0.39 \\ 
        Class 2 & 84.6±1.33 & 84.3±1.13 & 84.5±1.19 & 84.6±1.08 \\ 
        Class 3 & 84.6±1.28 & 85.9±0.93 & 85.9±1.01 & 86.0±0.63 \\ 
        Class 4 & 92.2±0.92 & 92.2±0.78 & 92.1±0.62 & 92.0±0.4 \\ 
        Class 5 & 87.7±1.7 & 87.3±1.94 & 87.6±1.63 & 87.5±1.69 \\ 
        Class 6 & 82.5±1.37 & 82.6±1.5 & 82.2±1.48 & 82.6±1.7 \\ 
        Class 7 & 72.4±1.99 & 72.4±1.39 & 72.5±1.73 & 73.2±1.6 \\ 
        Class 8 & 85.5±1.99 & 85.9±1.98 & 86.0±1.92 & 85.6±2.0 \\ 
        Class 9 & 99.4±0.98 & 99.5±0.88 & 99.4±0.93 & 99.3±0.95 \\ \bottomrule

\end{tabular}
}
\end{table}

\subsection{Evaluation on calibration metrics}
We have studied a total of four calibration metrics, including MinMax, Cosine, KL and MSE.
Results of different models on different datasets for different metrics are as follows:

\noindent\textbf{MinMax calibration.} Quantization scaling factors are computed based on the range directly determined by the maximum and minimum values of the feature map. 
The equation for calculating the scaling factors is as follows:
\begin{equation}
    s = \frac{max(x)-min(x)}{2^n-1}.
\end{equation}

\noindent\textbf{Cosine calibration.} 
Quantization range is determined based on the cosine distance between features before and after quantization.
Here we consider asymmetric quantization, where the minimum value after quantization is set to 0. 
Therefore, we only need to find the maximum value of the quantization range.
Search 100 times uniformly within the range of the maximum value of the feature to find the maximum value that minimizes cosine distance.
The equation is as follows:
\begin{equation}
    \min_sD_{cos}(x, x^q).
\end{equation}

\begin{table}[tb]
\centering
\caption{The influence of different metrics. We report the mean±std over 50 runs on CIFAR-10 dataset.}
\resizebox{\linewidth}{!}{
\begin{tabular}{@{}ccccc@{}}
\toprule
Dataset Size & Cosine         & KL        & MSE       & MinMax      \\ \midrule
\multicolumn{5}{c}{CIFAR-10 ResNet20 W4A4}                   \\ \midrule
        Average & 88.3±0.14 & 87.1±0.28 & 88.0±0.21 & 79.4±1.18 \\ 
        Class 0 & 92.4±0.54 & 92.9±0.54 & 92.2±0.44 & 88.3±1.46 \\ 
        Class 1 & 92.0±0.36 & 94.0±0.45 & 93.2±0.46 & 88.4±1.27 \\ 
        Class 2 & 83.5±0.65 & 83.6±0.62 & 84.6±0.62 & 70.7±1.94 \\ 
        Class 3 & 79.2±0.51 & 73.8±0.85 & 76.4±0.7 & 73.2±1.78 \\ 
        Class 4 & 86.9±0.48 & 84.5±0.77 & 86.6±0.61 & 71.0±2.99 \\ 
        Class 5 & 81.7±0.54 & 81.1±0.79 & 81.4±0.67 & 72.1±1.71 \\ 
        Class 6 & 94.3±0.33 & 93.8±0.46 & 94.2±0.45 & 89.4±1.06 \\ 
        Class 7 & 90.1±0.48 & 88.5±0.57 & 89.5±0.47 & 86.3±0.7 \\ 
        Class 8 & 89.6±0.48 & 85.9±0.87 & 88.8±0.71 & 64.9±3.47 \\ 
        Class 9 & 93.7±0.33 & 92.7±0.42 & 93.2±0.43 & 90.0±0.66 \\ \midrule
\multicolumn{5}{c}{CIFAR-10 ResNet32 W4A4}                \\ \midrule
        Average & 88.1±0.19 & 86.4±0.48 & 87.7±0.21 & 73.9±1.42 \\ 
        Class 0 & 92.0±0.48 & 90.9±1.0 & 91.8±0.54 & 85.9±2.42 \\ 
        Class 1 & 92.9±0.37 & 94.3±0.45 & 93.8±0.36 & 88.6±1.02 \\ 
        Class 2 & 85.2±0.58 & 85.3±0.85 & 86.1±0.63 & 72.2±2.55 \\ 
        Class 3 & 70.2±0.87 & 64.9±1.69 & 67.4±1.02 & 45.1±2.62 \\ 
        Class 4 & 91.6±0.48 & 86.7±1.24 & 90.3±0.49 & 72.5±2.3 \\ 
        Class 5 & 81.8±0.61 & 82.3±1.67 & 81.4±0.77 & 62.0±3.35 \\ 
        Class 6 & 92.1±0.45 & 89.0±1.16 & 92.3±0.54 & 88.8±2.23 \\ 
        Class 7 & 88.6±0.46 & 89.9±0.67 & 89.4±0.45 & 78.7±3.01 \\ 
        Class 8 & 94.2±0.28 & 90.1±1.63 & 93.2±0.52 & 63.5±4.22 \\ 
        Class 9 & 92.1±0.37 & 90.3±0.84 & 91.5±0.4 & 81.5±2.97 \\ \midrule
\multicolumn{5}{c}{CIFAR-10 ResNet44 W4A4}                \\ \midrule
        Average & 87.9±0.31 & 85.1±0.51 & 87.3±0.39 & 62.6±5.08 \\ 
        Class 0 & 78.5±1.09 & 83.2±1.51 & 77.4±1.13 & 27.3±8.72 \\ 
        Class 1 & 89.8±0.48 & 91.2±0.87 & 90.9±0.52 & 68.2±7.1 \\ 
        Class 2 & 84.6±0.86 & 80.4±1.32 & 83.9±0.97 & 56.3±6.9 \\ 
        Class 3 & 87.6±0.68 & 87.7±0.88 & 88.3±0.8 & 88.2±1.43 \\ 
        Class 4 & 91.8±0.44 & 87.9±1.21 & 91.2±0.55 & 73.4±5.74 \\ 
        Class 5 & 79.6±0.77 & 70.0±1.79 & 78.2±0.79 & 64.8±3.74 \\ 
        Class 6 & 90.6±0.83 & 85.5±1.73 & 87.7±1.16 & 61.2±11.26 \\ 
        Class 7 & 88.1±0.61 & 83.0±1.23 & 87.5±0.73 & 56.6±6.58 \\ 
        Class 8 & 94.4±0.51 & 91.4±0.82 & 94.2±0.52 & 54.2±10.08 \\ 
        Class 9 & 94.2±0.41 & 90.3±0.88 & 94.1±0.42 & 76.2±4.44 \\ \midrule
\multicolumn{5}{c}{CIFAR-10 ResNet56 W4A4}                \\ \midrule
        Average & 89.1±0.17 & 85.7±0.69 & 89.4±0.19 & 72.7±3.12 \\ 
        Class 0 & 84.9±0.54 & 88.5±1.09 & 86.8±0.82 & 72.6±8.02 \\ 
        Class 1 & 89.7±0.52 & 94.3±0.7 & 91.2±0.58 & 78.9±6.8 \\ 
        Class 2 & 84.1±0.57 & 84.3±1.12 & 85.2±0.68 & 59.5±6.61 \\ 
        Class 3 & 85.6±0.69 & 77.5±2.2 & 84.8±0.84 & 81.5±4.02 \\ 
        Class 4 & 92.0±0.48 & 73.7±3.08 & 91.6±0.61 & 51.5±8.77 \\ 
        Class 5 & 77.8±0.63 & 78.7±2.03 & 77.9±1.13 & 65.3±4.89 \\ 
        Class 6 & 94.8±0.33 & 90.8±1.08 & 94.7±0.52 & 80.0±4.4 \\ 
        Class 7 & 91.7±0.46 & 87.0±1.45 & 92.1±0.51 & 76.1±3.69 \\ 
        Class 8 & 96.9±0.24 & 90.4±1.36 & 96.4±0.31 & 75.4±6.05 \\ 
        Class 9 & 93.9±0.36 & 91.7±0.63 & 93.4±0.43 & 86.0±3.56 \\ \bottomrule
\end{tabular}
}
\end{table}
\begin{table}[htb]
\centering
\caption{The influence of different metrics. We report the mean±std over 50 runs on ImageNet dataset.}
\resizebox{\linewidth}{!}{
\begin{tabular}{@{}ccccc@{}}

\toprule
Dataset Size & Cosine         & KL        & MSE       & MinMax      \\ \midrule
\multicolumn{5}{c}{ImageNet ResNet18 W6A6}                   \\ \midrule
        Average & 70.2±0.05 & 70.3±0.04 & 70.3±0.06 & 69.8±0.15 \\ 
        Class 0 & 84.7±0.95 & 84.5±0.88 & 84.3±0.69 & 84.8±0.98 \\ 
        Class 1 & 89.6±0.83 & 89.3±0.96 & 89.5±0.85 & 89.6±1.42 \\ 
        Class 2 & 83.2±1.12 & 83.6±1.31 & 83.6±1.0 & 84.9±1.51 \\ 
        Class 3 & 68.2±2.33 & 68.9±2.05 & 68.6±1.85 & 65.8±3.65 \\ 
        Class 4 & 91.1±1.07 & 91.0±1.34 & 90.0±1.85 & 86.6±2.41 \\ 
        Class 5 & 70.4±1.4 & 69.9±0.69 & 70.9±1.14 & 70.4±1.92 \\ 
        Class 6 & 79.4±1.65 & 78.8±1.2 & 79.2±1.64 & 79.8±2.25 \\ 
        Class 7 & 69.4±1.34 & 68.8±2.15 & 68.5±2.26 & 68.4±2.2 \\ 
        Class 8 & 88.0±1.47 & 87.8±1.9 & 88.1±1.76 & 87.2±1.96 \\ 
        Class 9 & 98.7±0.95 & 100.0±0.0 & 100.0±0.0 & 99.6±0.8 \\ \midrule
\multicolumn{5}{c}{ImageNet ResNet50 W6A6}                \\ \midrule
        Average & 76.3±0.05 & 76.3±0.06 & 76.3±0.04 & 75.7±0.29 \\ 
        Class 0 & 94.2±0.6 & 94.3±0.8 & 95.0±1.08 & 93.6±1.91 \\ 
        Class 1 & 91.4±1.0 & 92.9±0.99 & 93.2±1.12 & 94.8±1.14 \\ 
        Class 2 & 81.9±0.56 & 81.4±0.9 & 81.7±0.9 & 82.9±2.12 \\ 
        Class 3 & 84.8±1.5 & 84.4±1.73 & 84.2±1.51 & 83.5±3.01 \\ 
        Class 4 & 87.7±0.73 & 87.5±0.88 & 86.3±1.44 & 83.4±2.67 \\ 
        Class 5 & 75.6±2.15 & 76.9±2.23 & 75.6±2.37 & 78.2±2.22 \\ 
        Class 6 & 81.8±1.73 & 80.8±1.69 & 82.0±1.83 & 79.2±2.85 \\ 
        Class 7 & 70.7±2.1 & 70.1±2.51 & 72.8±2.0 & 74.4±2.65 \\ 
        Class 8 & 84.0±0.85 & 84.1±1.35 & 84.0±0.94 & 83.3±2.55 \\ 
        Class 9 & 100.0±0.0 & 100.0±0.28 & 100.0±0.28 & 99.1±0.99 \\ \midrule
\multicolumn{5}{c}{ImageNet MobileNetV2 W6A6}                \\ \midrule
        Average & 70.1±0.06 & 69.9±0.07 & 70.2±0.06 & 69.7±0.07 \\ 
        Class 0 & 90.6±0.9 & 91.2±0.97 & 91.0±1.08 & 90.7±1.1 \\ 
        Class 1 & 85.2±1.14 & 85.2±1.05 & 85.3±0.95 & 86.6±1.22 \\ 
        Class 2 & 75.5±2.61 & 73.3±2.45 & 74.0±2.38 & 68.4±2.1 \\ 
        Class 3 & 78.2±1.94 & 78.8±1.92 & 78.0±1.83 & 79.2±2.0 \\ 
        Class 4 & 79.9±1.13 & 78.0±1.81 & 78.2±1.87 & 77.0±2.41 \\ 
        Class 5 & 70.0±2.4 & 65.9±1.76 & 68.4±2.01 & 67.5±2.82 \\ 
        Class 6 & 69.8±2.03 & 70.2±2.51 & 69.9±2.31 & 70.7±3.29 \\ 
        Class 7 & 73.8±1.28 & 73.3±1.48 & 73.1±1.39 & 73.7±1.62 \\ 
        Class 8 & 88.6±1.89 & 89.9±1.67 & 87.9±2.09 & 88.4±2.33 \\ 
        Class 9 & 97.8±0.65 & 97.7±0.69 & 97.3±1.03 & 97.4±0.93 \\ \midrule
\multicolumn{5}{c}{ImageNet MNasNet W6A6}                \\ \midrule
        Average & 74.9±0.09 & 74.9±0.26 & 74.8±0.19 & 72.9±1.41 \\ 
        Class 0 & 95.0±1.08 & 95.9±1.35 & 96.0±1.41 & 95.9±1.62 \\ 
        Class 1 & 91.2±1.26 & 91.0±1.56 & 91.2±1.59 & 90.6±2.53 \\ 
        Class 2 & 82.2±2.16 & 81.8±2.15 & 79.5±2.87 & 74.6±5.24 \\ 
        Class 3 & 81.5±2.76 & 82.8±2.49 & 82.4±2.95 & 79.2±5.04 \\ 
        Class 4 & 81.4±2.33 & 81.0±1.84 & 82.0±2.21 & 75.8±4.79 \\ 
        Class 5 & 78.6±2.23 & 76.8±2.27 & 79.1±2.01 & 72.1±3.45 \\ 
        Class 6 & 80.3±2.15 & 82.6±1.92 & 79.9±2.65 & 77.9±3.33 \\ 
        Class 7 & 65.6±2.65 & 66.4±1.7 & 67.7±3.5 & 63.2±6.37 \\ 
        Class 8 & 88.5±1.72 & 90.1±1.83 & 88.9±2.3 & 89.2±3.92 \\ 
        Class 9 & 97.1±1.27 & 97.6±1.34 & 97.6±1.13 & 97.1±1.39 \\ 
        \bottomrule
\end{tabular}
}
\end{table}

\begin{table}[htb]
\centering
\caption{The influence of different metrics. We report the mean±std over 50 runs on ImageNet dataset.}
\resizebox{\linewidth}{!}{
\begin{tabular}{@{}ccccc@{}}

\toprule
Dataset Size & Cosine         & KL        & MSE       & MinMax      \\ \midrule
\multicolumn{5}{c}{ImageNet RegNetx600M W6A6}                   \\ \midrule
        Average & 72.9±0.06 & 73.0±0.06 & 73.0±0.08 & 72.4±0.11 \\ 
        Class 0 & 90.2±0.72 & 90.6±1.16 & 90.4±0.77 & 90.3±1.46 \\ 
        Class 1 & 89.6±0.8 & 89.5±0.88 & 89.3±1.11 & 89.4±1.7 \\ 
        Class 2 & 76.2±1.59 & 75.8±1.54 & 75.8±1.86 & 76.0±2.56 \\ 
        Class 3 & 76.5±2.11 & 75.9±1.46 & 77.0±1.84 & 76.5±2.45 \\ 
        Class 4 & 90.8±1.64 & 90.3±1.74 & 89.9±1.49 & 88.2±2.67 \\ 
        Class 5 & 77.1±1.66 & 76.2±1.91 & 75.8±1.99 & 75.0±2.27 \\ 
        Class 6 & 75.6±2.63 & 77.4±2.31 & 77.0±2.37 & 76.7±2.67 \\ 
        Class 7 & 75.2±1.79 & 75.2±2.37 & 75.0±2.01 & 74.3±2.33 \\ 
        Class 8 & 83.0±1.84 & 83.9±1.32 & 83.4±1.79 & 82.3±2.28 \\ 
        Class 9 & 96.0±0.0 & 95.9±0.47 & 96.0±0.28 & 95.8±0.78 \\ \midrule
\multicolumn{5}{c}{ImageNet RegNetx3200M W6A6}                \\ \midrule
        Average & 77.9±0.04 & 77.9±0.06 & 77.9±0.07 & 77.1±0.14 \\ 
        Class 0 & 94.0±0.28 & 94.0±0.0 & 94.0±0.28 & 95.0±1.15 \\ 
        Class 1 & 90.8±1.2 & 91.4±1.02 & 91.8±0.65 & 92.4±1.74 \\ 
        Class 2 & 85.1±0.99 & 84.3±0.98 & 84.5±1.19 & 82.0±1.65 \\ 
        Class 3 & 84.6±1.29 & 85.4±0.93 & 85.9±1.01 & 85.0±2.09 \\ 
        Class 4 & 92.4±0.77 & 92.3±1.2 & 92.1±0.62 & 91.5±1.32 \\ 
        Class 5 & 87.7±0.93 & 89.1±1.34 & 87.6±1.63 & 86.9±2.08 \\ 
        Class 6 & 82.9±1.34 & 82.9±1.45 & 82.2±1.48 & 79.2±3.73 \\ 
        Class 7 & 74.0±1.52 & 70.8±1.6 & 72.5±1.73 & 73.0±1.84 \\ 
        Class 8 & 84.8±1.75 & 85.1±1.84 & 86.0±1.92 & 85.4±1.98 \\ 
        Class 9 & 100.0±0.0 & 100.0±0.28 & 99.4±0.93 & 97.8±0.67 \\ \bottomrule

\end{tabular}
}
\end{table}
\noindent\textbf{KL calibration.} 
Minimizing the KL divergence between the distributions before and after quantization to find the range for quantization. Specifically, the distribution is divided into a histogram of 2048 bins, and the difference between the distributions before and after quantization is compared. The equation is as follows:
\begin{equation}
    \min_sD_{cos}(hist(x), hist(x^q)).
\end{equation}

\noindent\textbf{MSE calibration.} 
Similar to KL calibration, the quantization range is determined based on the difference between the pre- and post-quantization distributions using MSE distance instead of KL divergence. Similarly, search 100 times to find the optimal maximum value. The equation is as follows:
\begin{equation}
    \min_s\|x - x^q\|^2.
\end{equation}

\subsection{Evaluation on bitwidth}
We evaluated the performance of multiple models under 6-bit and 8-bit quantization for the ImageNet dataset, and under 4-bit and 6-bit quantization for the CIFAR-10 dataset.
The experimental results are shown as follows:

\begin{table}[htb]
\centering
\caption{The influence of quantization settings on CIFAR-10 dataset.}
\label{table_quantization_settings}
\resizebox{\linewidth}{!}{
\begin{tabular}{@{}ccccc@{}}
\toprule
Task     & \multicolumn{2}{c}{CIFAR-10 ResNet20} & \multicolumn{2}{c}{CIFAR-10 ResNet32} \\
bit-width & W6A6              & W4A4              & W6A6              & W4A4               \\ \midrule
        Average & 90.7±0.08 & 88.0±0.21 & 90.4±0.1 & 87.7±0.21 \\ 
        Class 0 & 90.1±0.27 & 92.2±0.44 & 93.5±0.3 & 91.8±0.54 \\ 
        Class 1 & 95.8±0.19 & 93.2±0.46 & 95.8±0.22 & 93.8±0.36 \\ 
        Class 2 & 88.5±0.3 & 84.6±0.62 & 90.8±0.3 & 86.1±0.63 \\ 
        Class 3 & 82.6±0.51 & 76.4±0.7 & 77.4±0.41 & 67.4±1.02 \\ 
        Class 4 & 90.8±0.27 & 86.6±0.61 & 90.0±0.35 & 90.3±0.49 \\ 
        Class 5 & 85.5±0.33 & 81.4±0.67 & 84.4±0.35 & 81.4±0.77 \\ 
        Class 6 & 92.1±0.28 & 94.2±0.45 & 94.6±0.3 & 92.3±0.54 \\ 
        Class 7 & 93.2±0.29 & 89.5±0.47 & 89.5±0.26 & 89.4±0.45 \\ 
        Class 8 & 93.1±0.21 & 88.8±0.71 & 94.3±0.29 & 93.2±0.52 \\ 
        Class 9 & 94.8±0.21 & 93.2±0.43 & 93.6±0.2 & 91.5±0.4 \\ \bottomrule
\end{tabular}
}
\end{table}

\begin{table}[htb]
\centering
\caption{The influence of quantization settings on CIFAR-10 dataset.}
\label{table_quantization_settings}
\resizebox{\linewidth}{!}{
\begin{tabular}{@{}ccccc@{}}
\toprule
Task     & \multicolumn{2}{c}{CIFAR-10 ResNet44} & \multicolumn{2}{c}{CIFAR-10 ResNet56} \\
bit-width & W6A6              & W4A4              & W6A6              & W4A4               \\ \midrule
        Average & 92.1±0.1 & 87.3±0.39 & 92.9±0.11 & 89.4±0.19 \\ 
        Class 0 & 92.0±0.28 & 77.4±1.13 & 93.1±0.28 & 86.8±0.82 \\ 
        Class 1 & 96.7±0.2 & 90.9±0.52 & 96.6±0.22 & 91.2±0.58 \\ 
        Class 2 & 90.8±0.31 & 83.9±0.97 & 90.6±0.39 & 85.2±0.68 \\ 
        Class 3 & 86.4±0.35 & 88.3±0.8 & 86.5±0.36 & 84.8±0.84 \\ 
        Class 4 & 92.5±0.25 & 91.2±0.55 & 94.2±0.33 & 91.6±0.61 \\ 
        Class 5 & 86.9±0.35 & 78.2±0.79 & 87.9±0.42 & 77.9±1.13 \\ 
        Class 6 & 92.0±0.36 & 87.7±1.16 & 94.8±0.21 & 94.7±0.52 \\ 
        Class 7 & 93.9±0.26 & 87.5±0.73 & 94.1±0.31 & 92.1±0.51 \\ 
        Class 8 & 95.3±0.26 & 94.2±0.52 & 95.5±0.22 & 96.4±0.31 \\ 
        Class 9 & 94.6±0.26 & 94.1±0.42 & 95.7±0.21 & 93.4±0.43 \\ \bottomrule
\end{tabular}
}
\end{table}

\begin{table}[htb]
\centering
\caption{The influence of quantization settings on Imagenet dataset.}
\label{table_quantization_settings}
\resizebox{\linewidth}{!}{
\begin{tabular}{@{}ccccc@{}}
\toprule
Task     & \multicolumn{2}{c}{ImageNet ResNet18} & \multicolumn{2}{c}{Imagenet ResNet50} \\
bit-width & W8A8                & W6A6              & W8A8                & W6A6               \\ \midrule
        Average & 70.9±0.03 & 70.3±0.06 & 76.6±0.03 & 76.3±0.04 \\ 
        Class 0 & 86.0±0.0 & 84.3±0.69 & 93.4±1.47 & 95.0±1.08 \\ 
        Class 1 & 86.8±0.97 & 89.5±0.85 & 93.8±0.54 & 93.2±1.12 \\ 
        Class 2 & 78.6±1.29 & 83.6±1.0 & 82.0±0.0 & 81.7±0.9 \\ 
        Class 3 & 69.1±1.07 & 68.6±1.85 & 83.8±0.65 & 84.2±1.51 \\ 
        Class 4 & 88.0±0.0 & 90.0±1.85 & 85.0±1.08 & 86.3±1.44 \\ 
        Class 5 & 72.0±0.0 & 70.9±1.14 & 78.8±1.39 & 75.6±2.37 \\ 
        Class 6 & 74.6±1.23 & 79.2±1.64 & 75.2±1.45 & 82.0±1.83 \\ 
        Class 7 & 68.1±0.39 & 68.5±2.26 & 71.5±1.99 & 72.8±2.0 \\ 
        Class 8 & 86.8±1.05 & 88.1±1.76 & 82.9±1.21 & 84.0±0.94 \\ 
        Class 9 & 100.0±0.0 & 100.0±0.0 & 100.0±0.0 & 100.0±0.28 \\ \midrule
Task     & \multicolumn{2}{c}{ImageNet MobilenetV2} & \multicolumn{2}{c}{Imagenet MNasNet} \\
bit-width & W8A8                & W6A6              & W8A8                & W6A6               \\ \midrule
        Average & 72.0±0.03 & 70.2±0.06 & 76.4±0.04 & 74.8±0.19 \\ 
        Class 0 & 91.9±0.47 & 91.0±1.08 & 97.0±1.0 & 96.0±1.41 \\ 
        Class 1 & 92.0±0.0 & 85.3±0.95 & 89.6±0.87 & 91.2±1.59 \\ 
        Class 2 & 88.8±1.07 & 74.0±2.38 & 86.9±1.39 & 79.5±2.87 \\ 
        Class 3 & 79.2±1.05 & 78.0±1.83 & 81.5±2.02 & 82.4±2.95 \\ 
        Class 4 & 83.0±1.0 & 78.2±1.87 & 84.4±1.68 & 82.0±2.21 \\ 
        Class 5 & 68.6±1.66 & 68.4±2.01 & 76.3±1.27 & 79.1±2.01 \\ 
        Class 6 & 78.8±1.27 & 69.9±2.31 & 75.9±1.67 & 79.9±2.65 \\ 
        Class 7 & 77.0±1.22 & 73.1±1.39 & 70.8±1.33 & 67.7±3.5 \\ 
        Class 8 & 86.0±0.0 & 87.9±2.09 & 90.0±0.0 & 88.9±2.3 \\ 
        Class 9 & 96.1±0.39 & 97.3±1.03 & 96.0±0.0 & 97.6±1.13 \\ \bottomrule
\end{tabular}
}
\end{table}

\begin{table}[htb]
\centering
\caption{The influence of quantization settings on Imagenet dataset.}
\label{table_quantization_settings}
\resizebox{\linewidth}{!}{
\begin{tabular}{@{}ccccc@{}}
\toprule
Task     & \multicolumn{2}{c}{ImageNet RegNet600} & \multicolumn{2}{c}{ImageNet RegNet3200} \\
bit-width & W8A8                & W6A6              & W8A8                & W6A6               \\ \midrule
        Average & 73.5±0.04 & 73.0±0.08 & 78.5±0.03 & 77.9±0.07 \\ 
        Class 0 & 92.0±0.0 & 90.4±0.77 & 94.0±0.0 & 94.0±0.28 \\ 
        Class 1 & 88.0±0.85 & 89.3±1.11 & 90.0±0.28 & 91.8±0.65 \\ 
        Class 2 & 76.9±1.15 & 75.8±1.86 & 85.8±0.65 & 84.5±1.19 \\ 
        Class 3 & 78.1±0.84 & 77.0±1.84 & 84.0±0.28 & 85.9±1.01 \\ 
        Class 4 & 85.2±1.74 & 89.9±1.49 & 92.0±0.28 & 92.1±0.62 \\ 
        Class 5 & 75.5±0.94 & 75.8±1.99 & 85.5±1.25 & 87.6±1.63 \\ 
        Class 6 & 79.6±1.25 & 77.0±2.37 & 84.8±0.98 & 82.2±1.48 \\ 
        Class 7 & 71.7±1.55 & 75.0±2.01 & 72.9±1.0 & 72.5±1.73 \\ 
        Class 8 & 88.0±0.0 & 83.4±1.79 & 86.1±1.6 & 86.0±1.92 \\ 
        Class 9 & 96.0±0.0 & 96.0±0.28 & 100.0±0.0 & 99.4±0.93 \\ \bottomrule
\end{tabular}
}
\end{table}

\subsection{Evaluation on noise data}
We investigated the performance of quantized models when introducing noise data similar to the actual calibration set during calibration. 
Specifically, we conducted experiments on the performance of quantized models when the calibration set contains 1\%, 5\%, 10\%, and 50\% noise data. 
The results of the experiments on multiple models are as follows, we only plot 10 classes of ImageNet.
The figures below are the relative performance change to the clean case with varying datanoise amounts.
The change values are demonstrated in different colors.

\begin{figure}[htb]
    \centering
    \includegraphics[width=\linewidth]{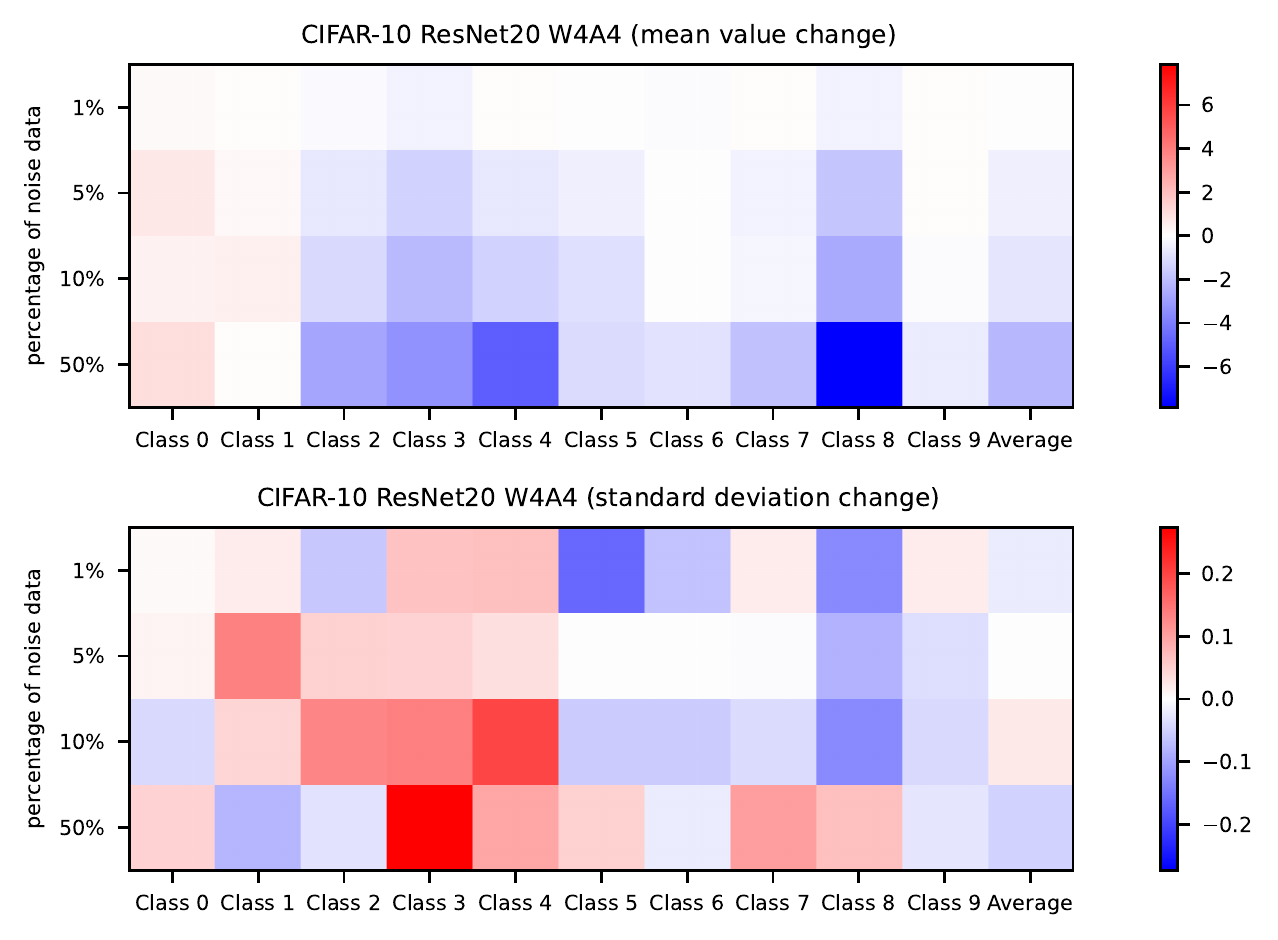} 
\end{figure}
\vspace{-1cm}
\begin{figure}[htb]
    \centering
    \includegraphics[width=\linewidth]{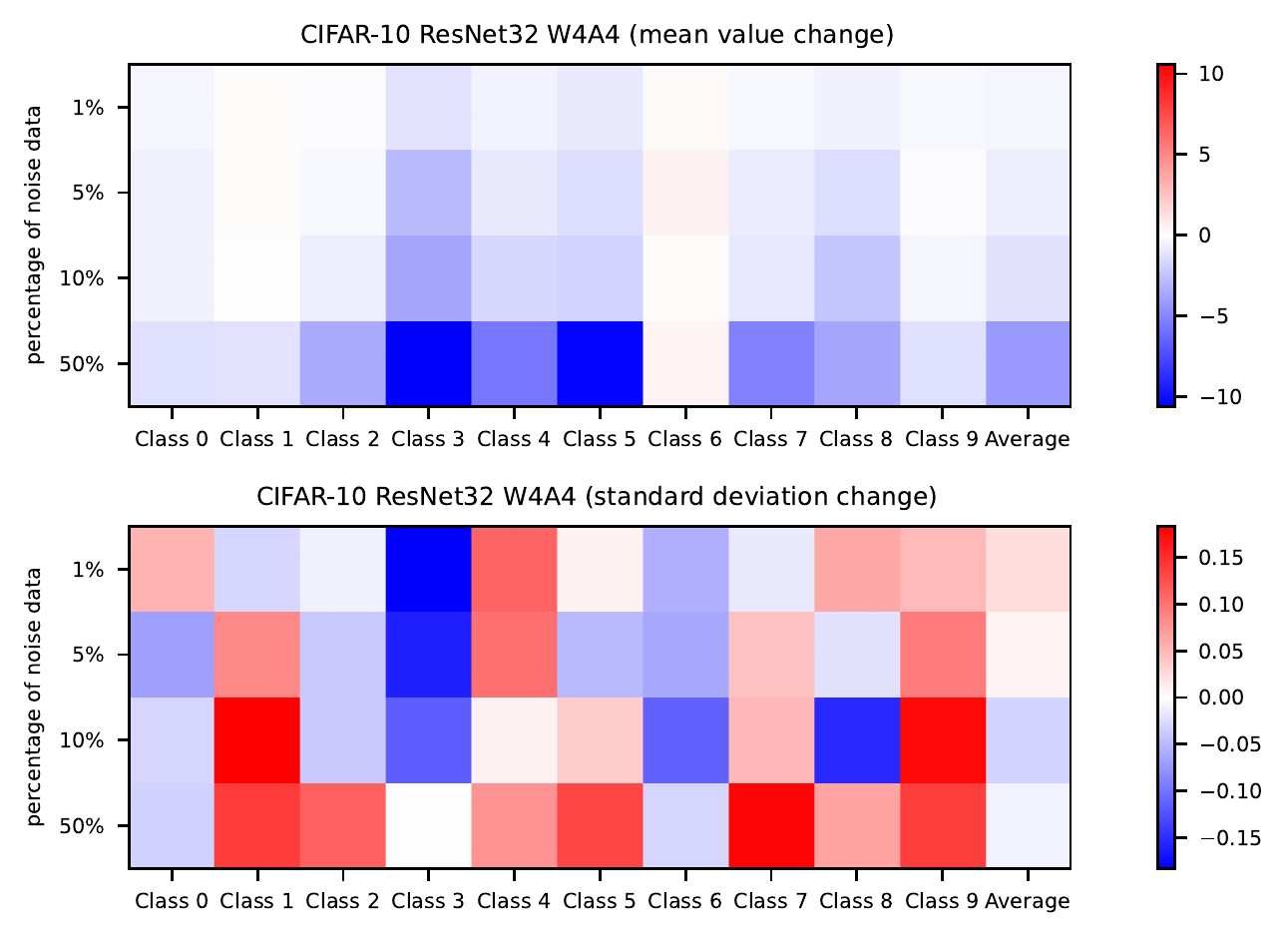}
\end{figure}
\vspace{-1cm}
\begin{figure}[H]
    \centering
    \includegraphics[width=\linewidth]{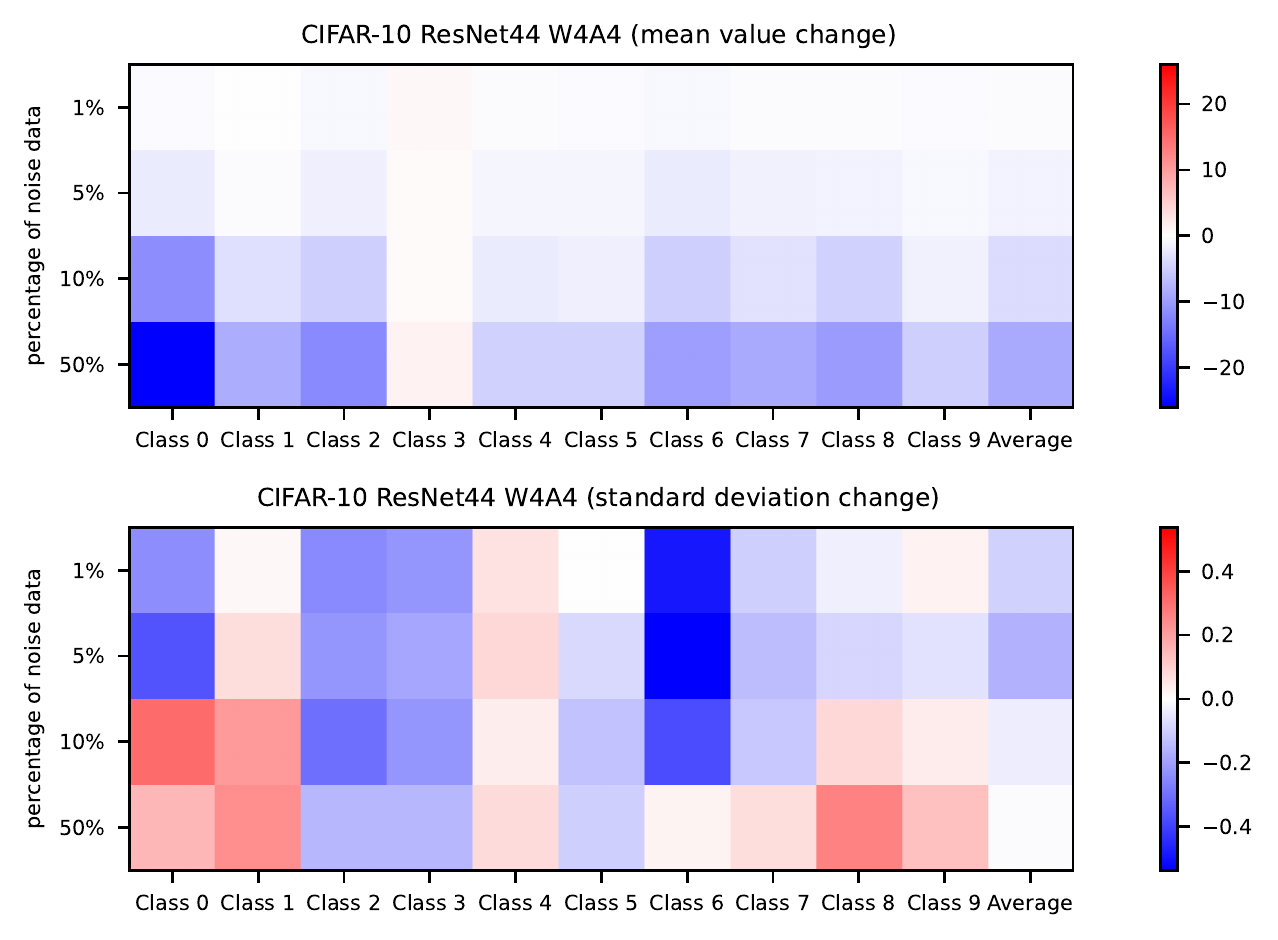} 
\end{figure}
\vspace{-1cm}
\begin{figure}[H]
    \centering
    \includegraphics[width=\linewidth]{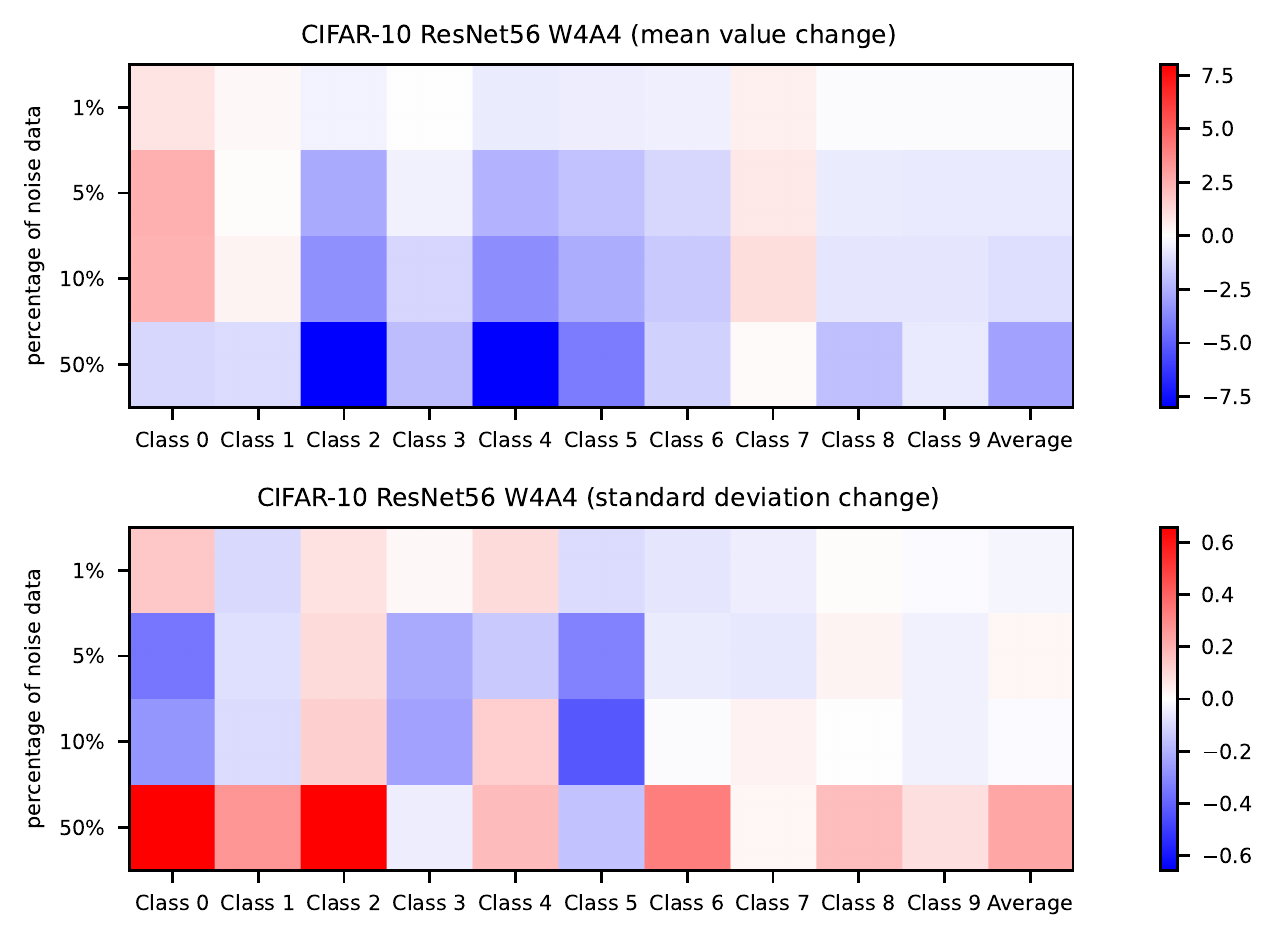} 
\end{figure}
\vspace{-1cm}
\begin{figure}[H]
    \centering
    \includegraphics[width=\linewidth]{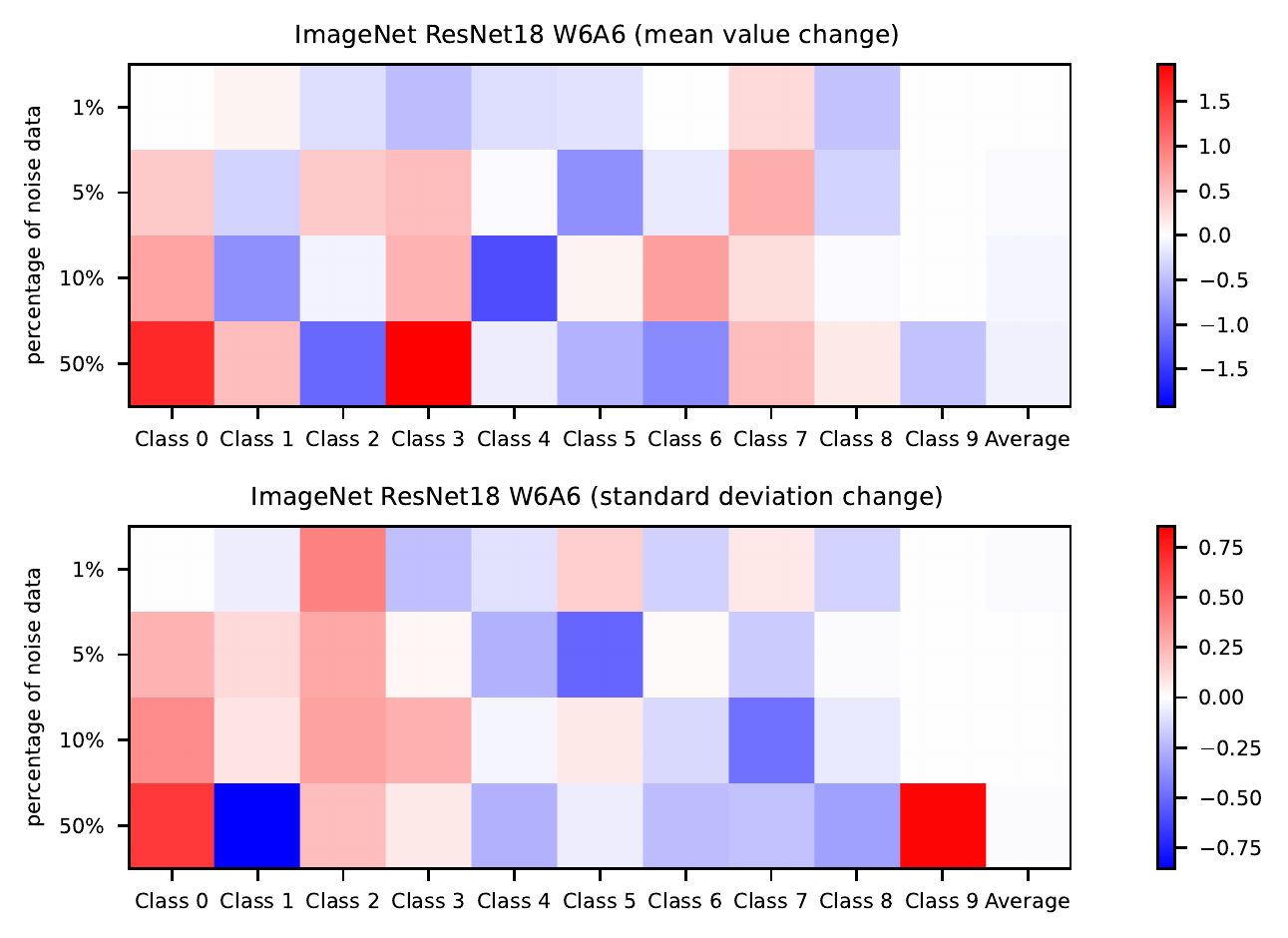} 
\end{figure}
\vspace{-1cm}
\begin{figure}[H]
    \centering
    \includegraphics[width=\linewidth]{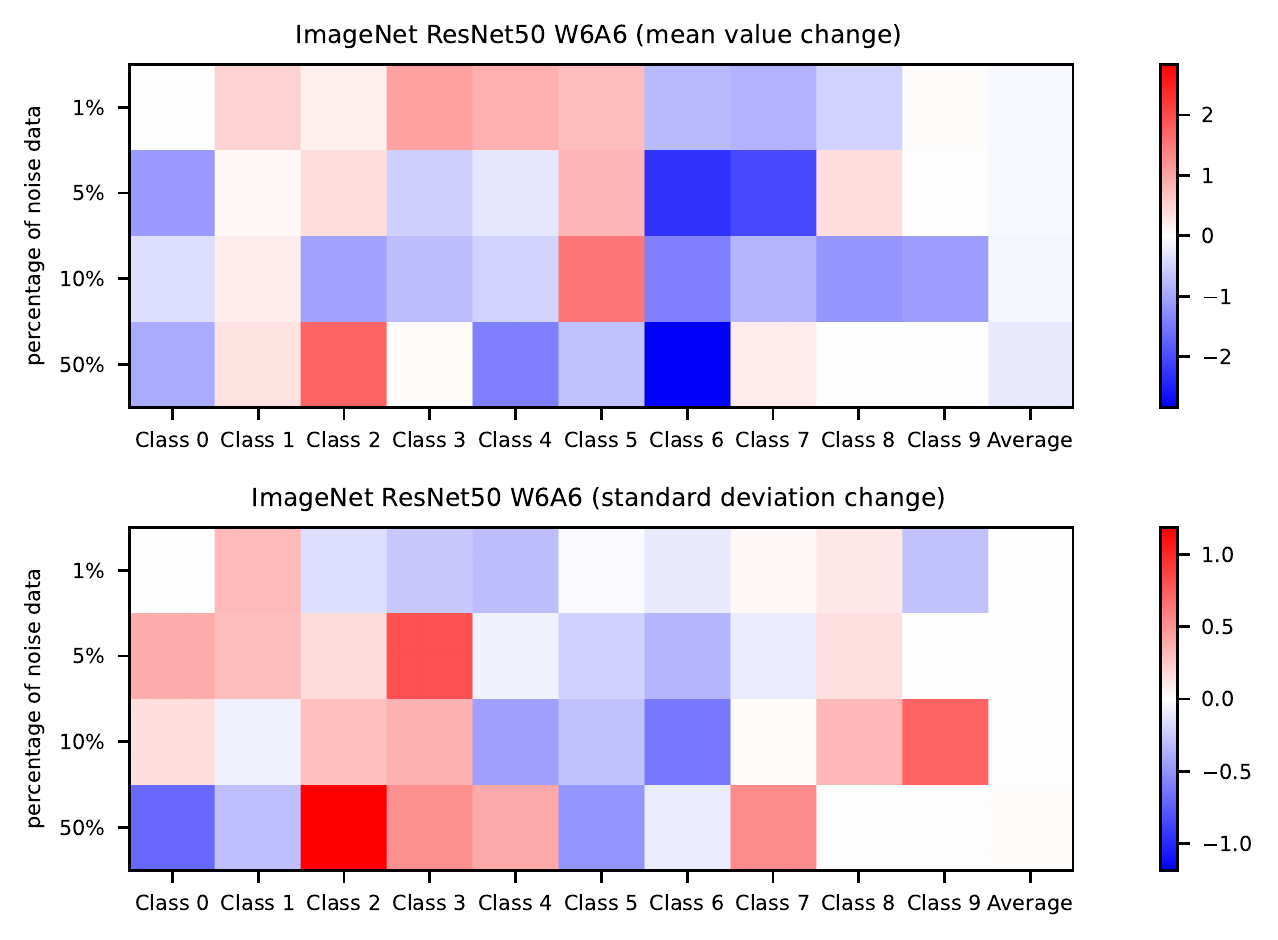}
\end{figure}
\vspace{-1cm}
\begin{figure}[H]
    \centering
    \includegraphics[width=\linewidth]{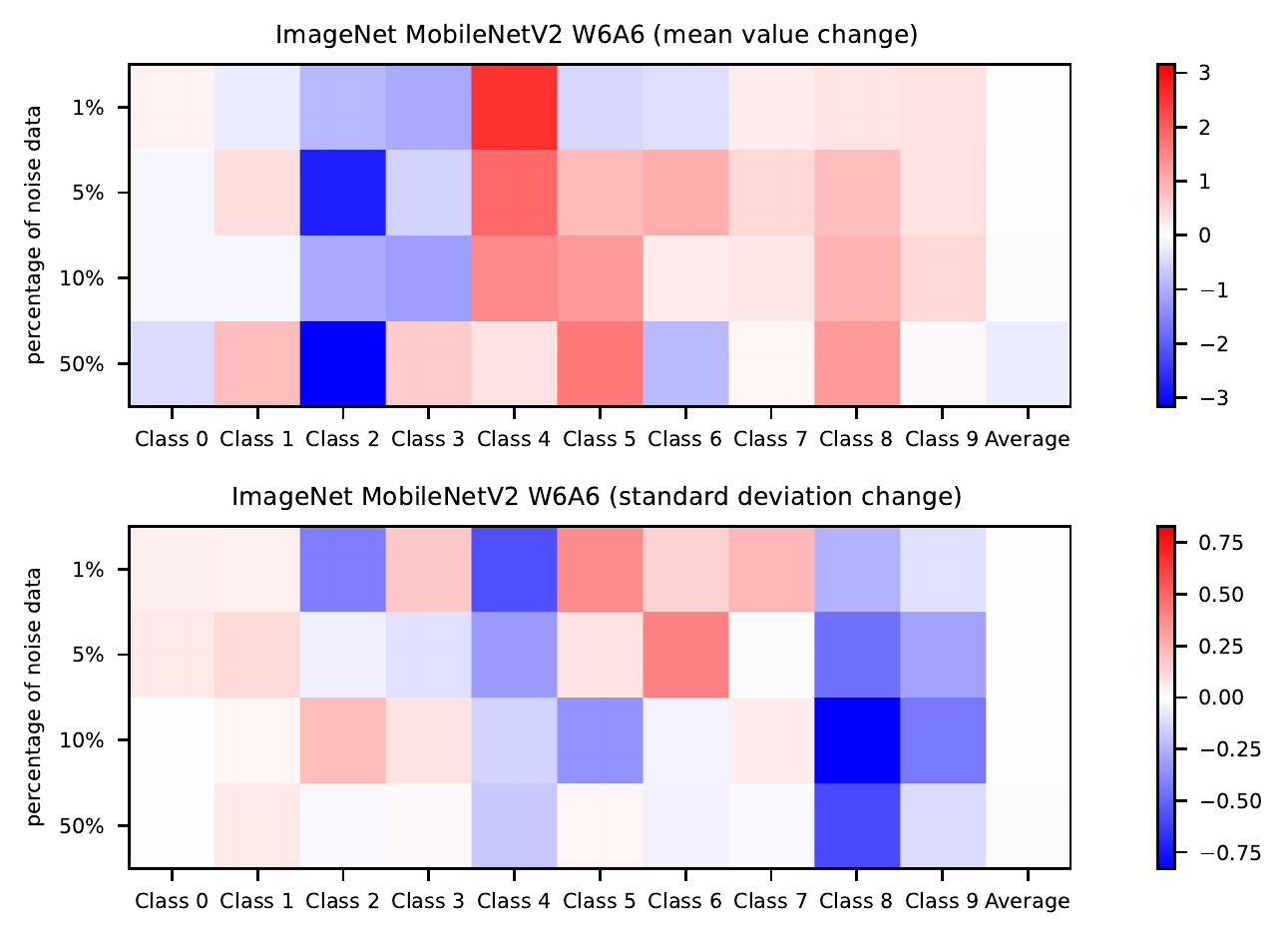}
\end{figure}
\vspace{-1cm}
\begin{figure}[H]
    \centering
    \includegraphics[width=\linewidth]{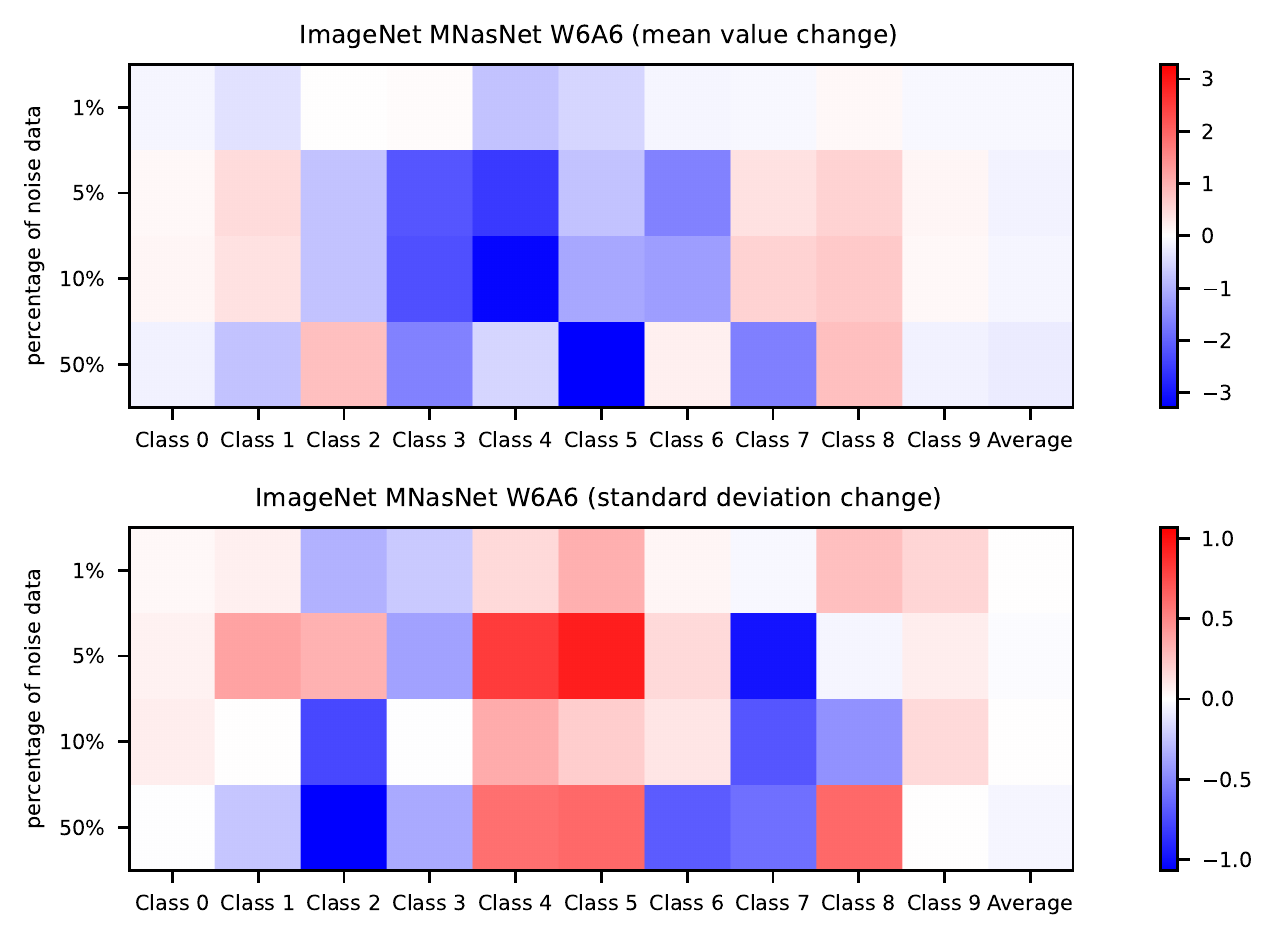}
\end{figure}
\vspace{-1cm}
\begin{figure}[H]
    \centering
    \includegraphics[width=\linewidth]{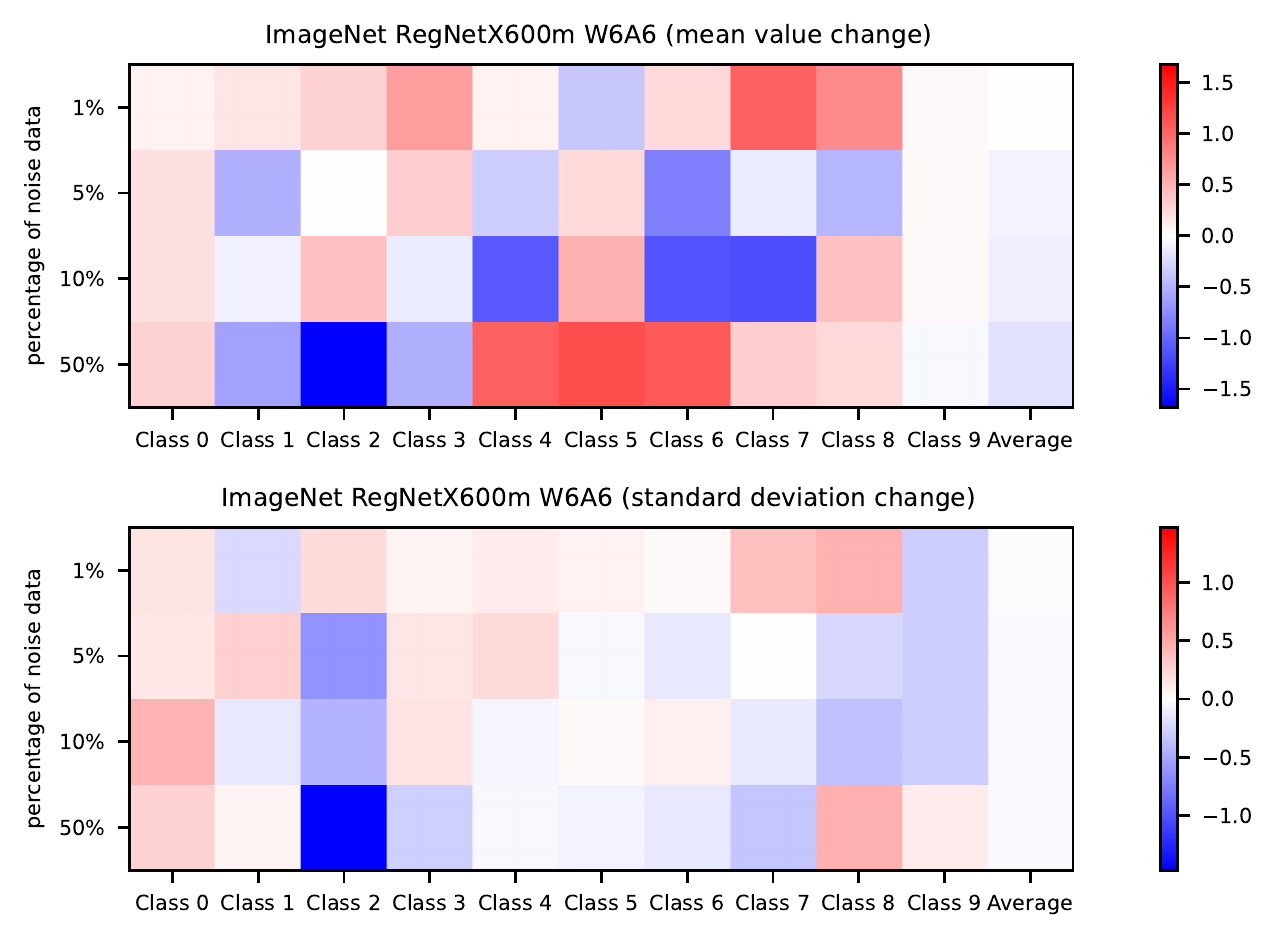}
\end{figure}
\vspace{-1cm}
\begin{figure}[H]
    \centering
    \includegraphics[width=\linewidth]{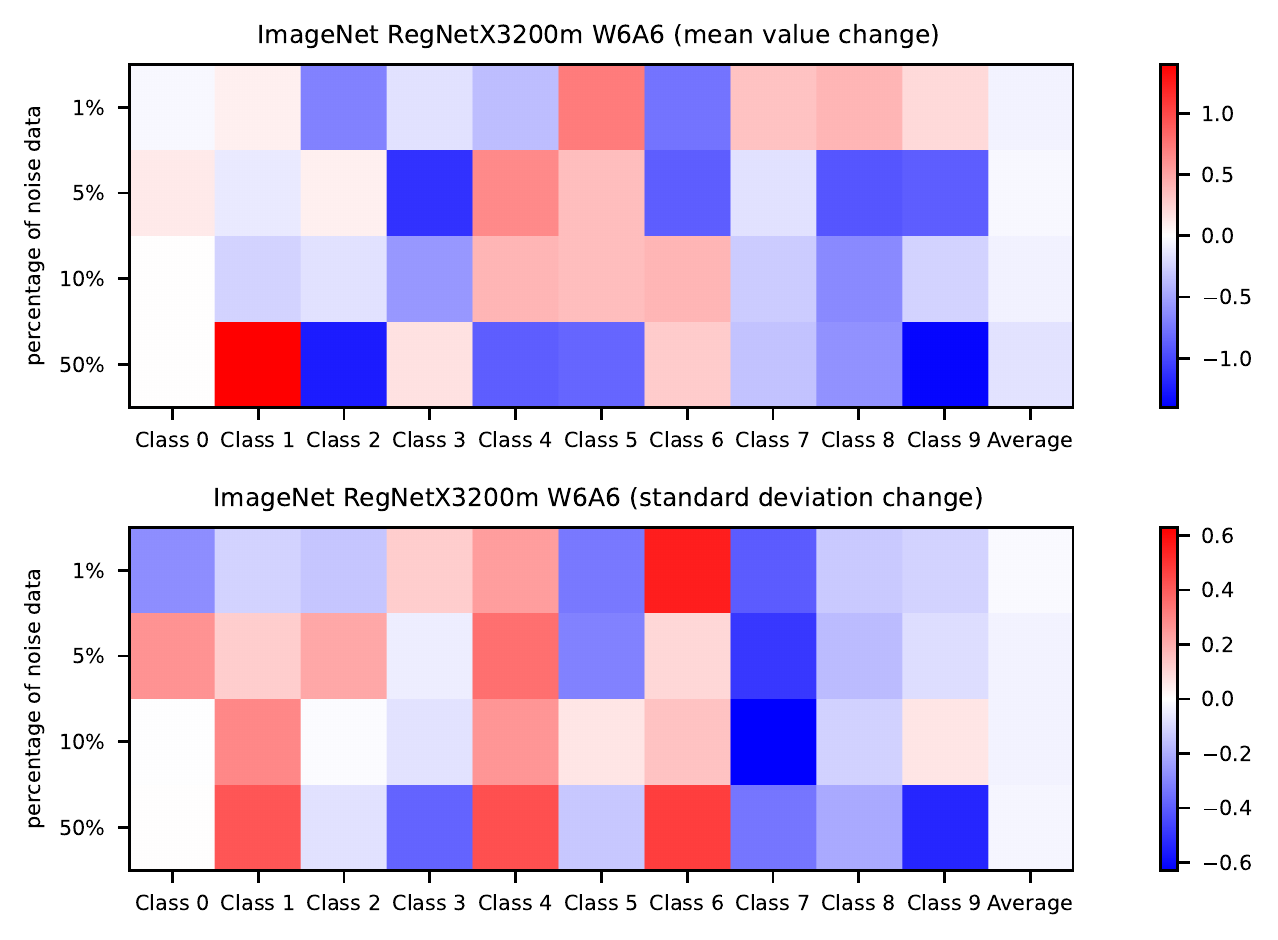}
\end{figure}

\subsection{Evaluation on class bias}
We conducted experiments on unbalanced calibration datasets to explore how class imbalance affects model quantization. 
Specifically, we set one category in the calibration dataset to represent 50\% of the total samples and tested four different categories across multiple models.
The experimental results of top-1 accuracy is averaged over 50 runs with different random seeds. The prediction accuracy change is demonstrated in different colors (red means increment and blue means decrement).
We only plot 10 classes of ImageNet.

\begin{figure}[htb]
    \centering
    \includegraphics[width=\linewidth]{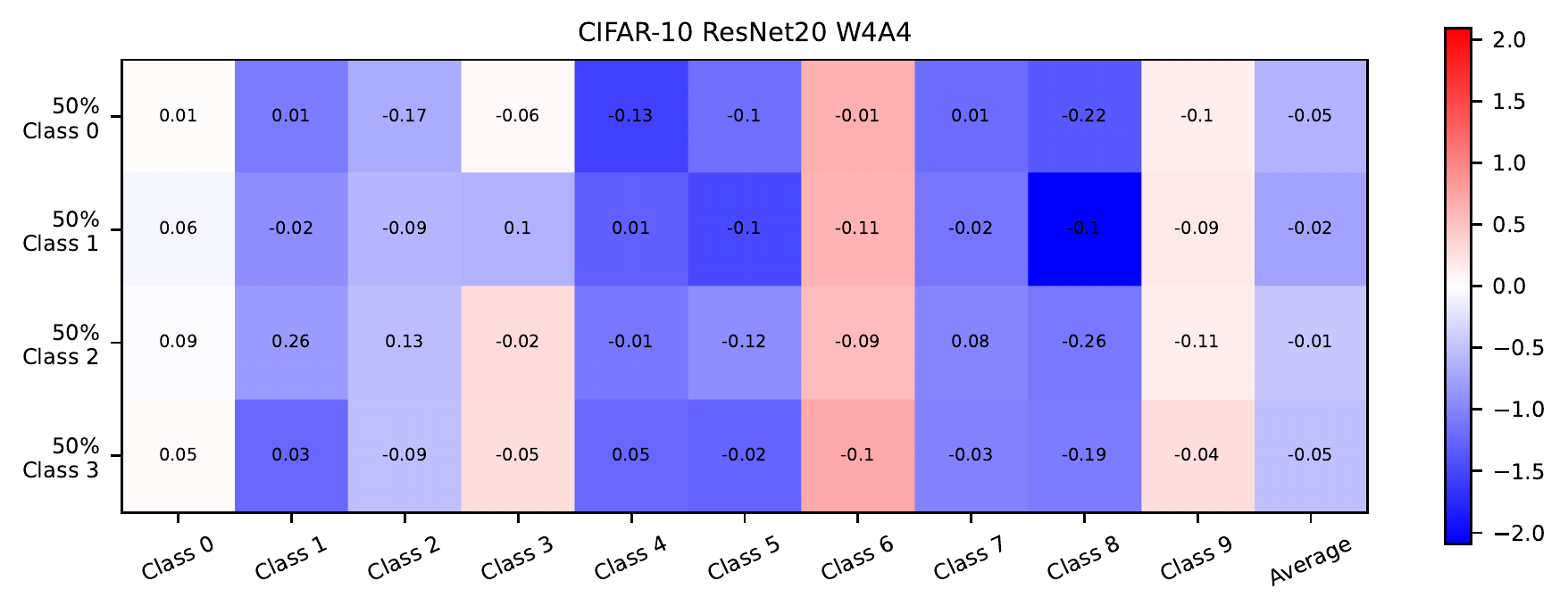} 
\end{figure}
\vspace{-1cm}
\begin{figure}[htb]
    \centering
    \includegraphics[width=\linewidth]{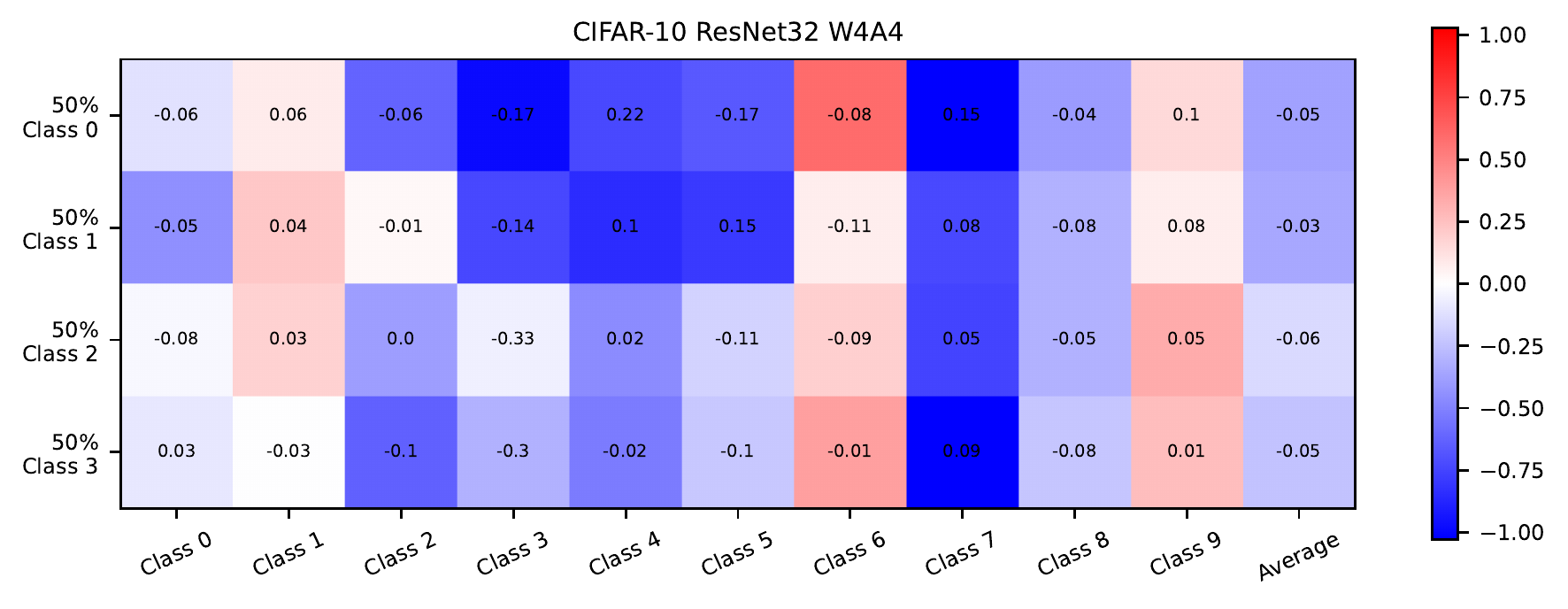}
\end{figure}
\vspace{-1cm}
\begin{figure}[H]
    \centering
    \includegraphics[width=\linewidth]{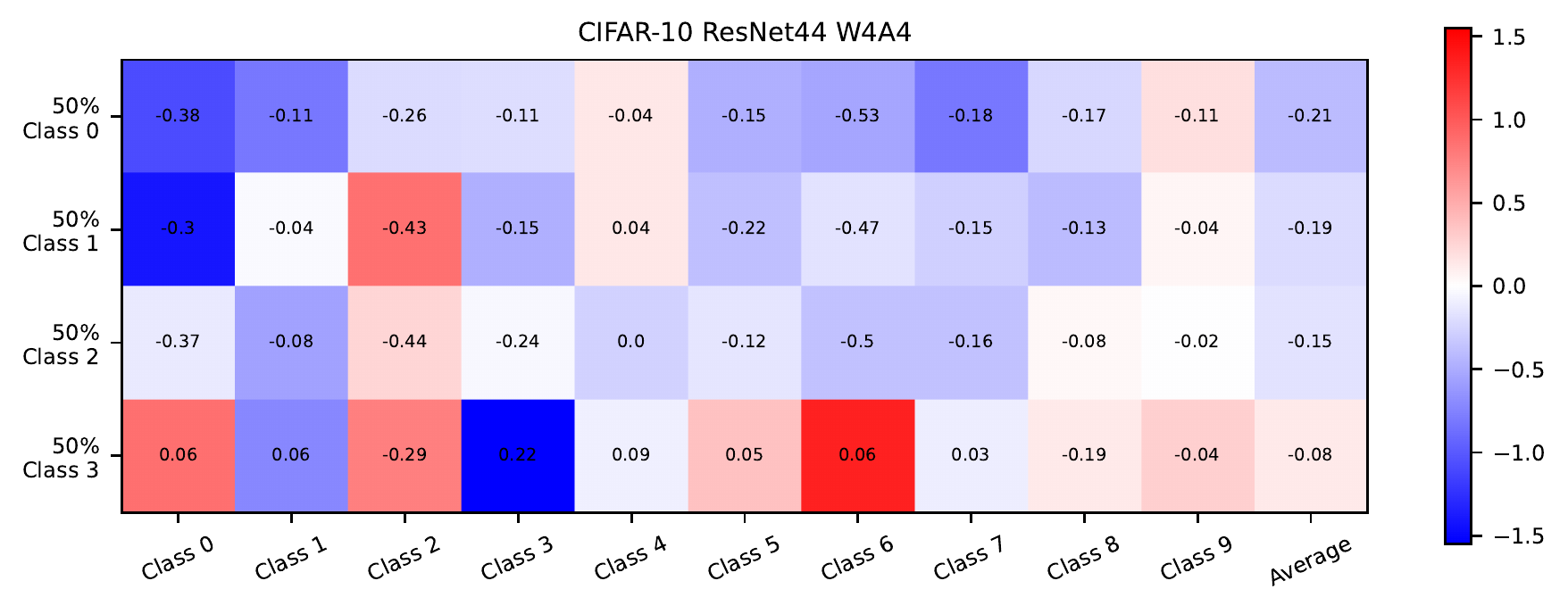} 
\end{figure}
\vspace{-1cm}
\begin{figure}[H]
    \centering
    \includegraphics[width=\linewidth]{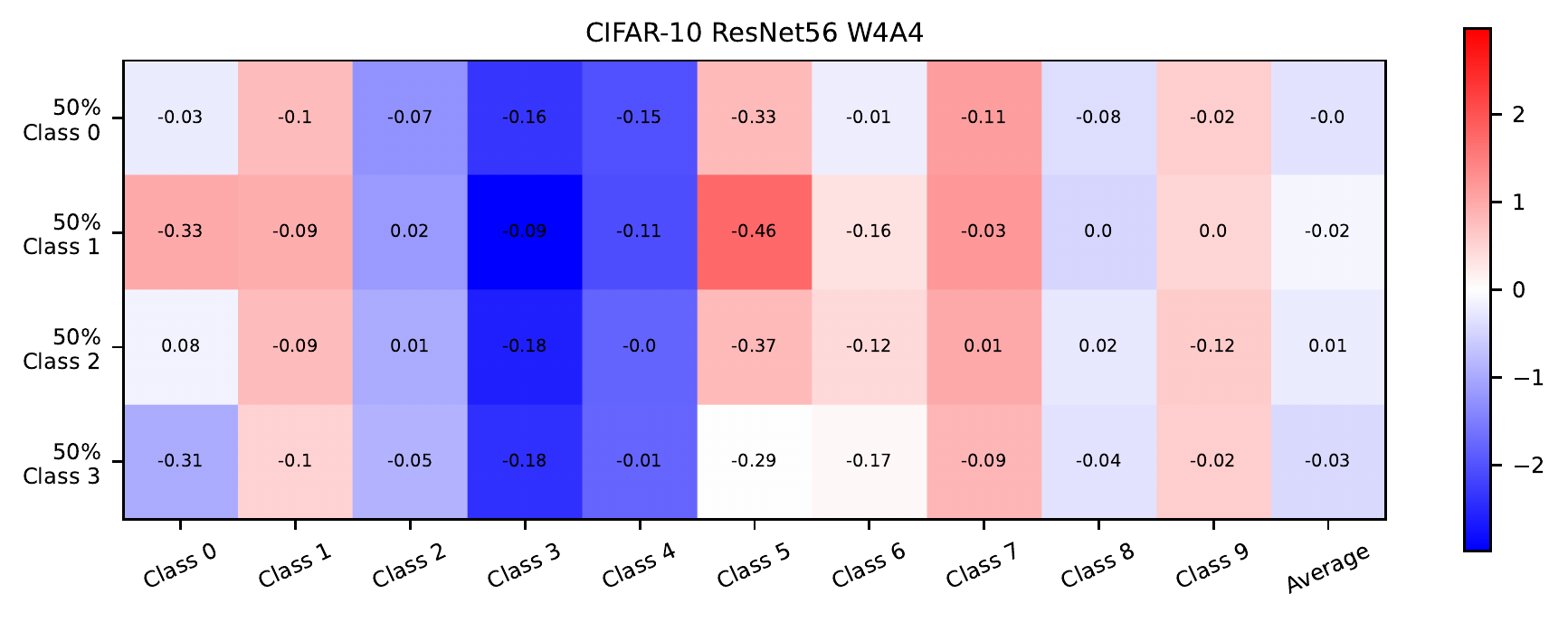} 
\end{figure}
\vspace{-1cm}
\begin{figure}[H]
    \centering
    \includegraphics[width=\linewidth]{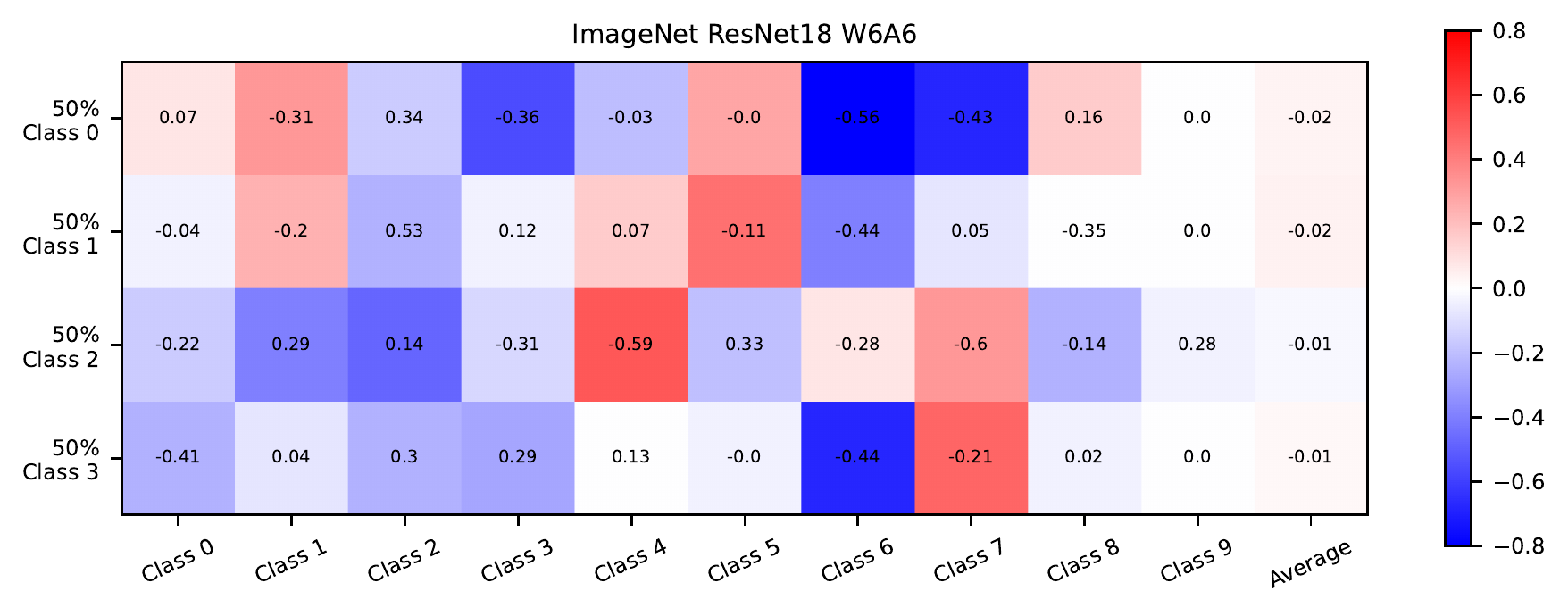} 
\end{figure}
\vspace{-1cm}
\begin{figure}[H]
    \centering
    \includegraphics[width=\linewidth]{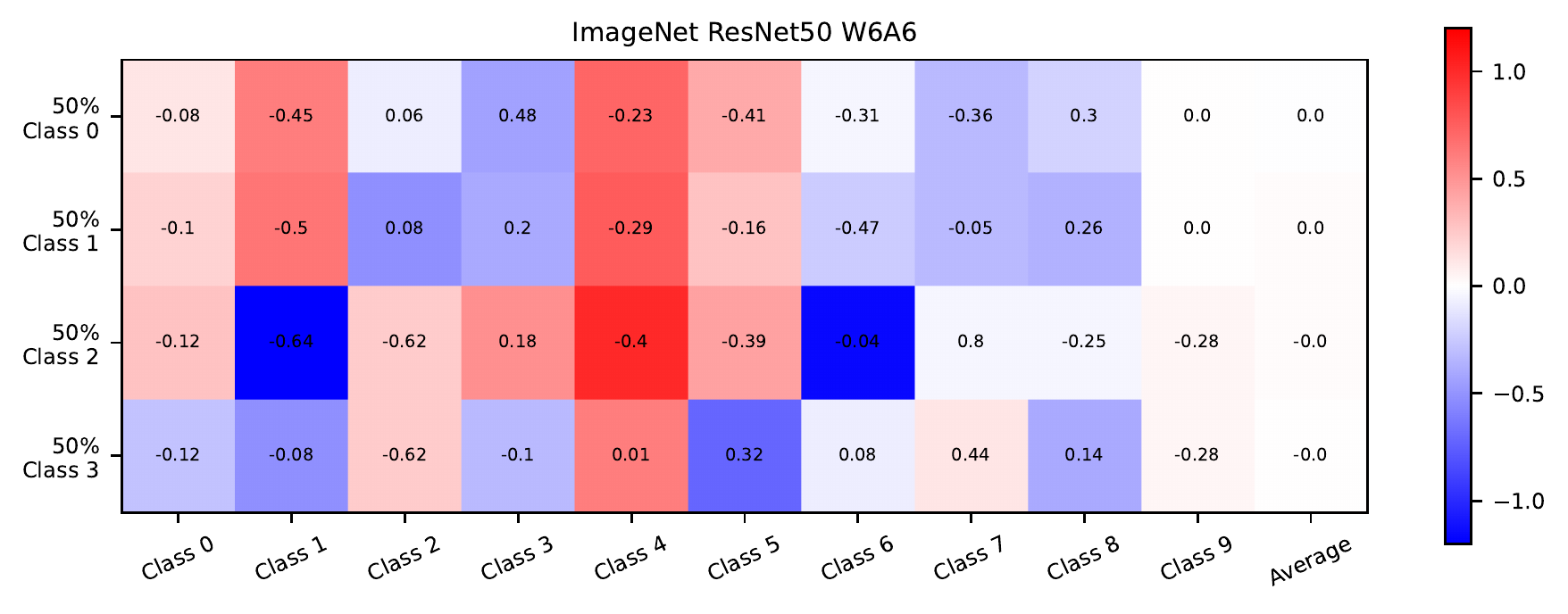}
\end{figure}
\vspace{-1cm}
\begin{figure}[H]
    \centering
    \includegraphics[width=\linewidth]{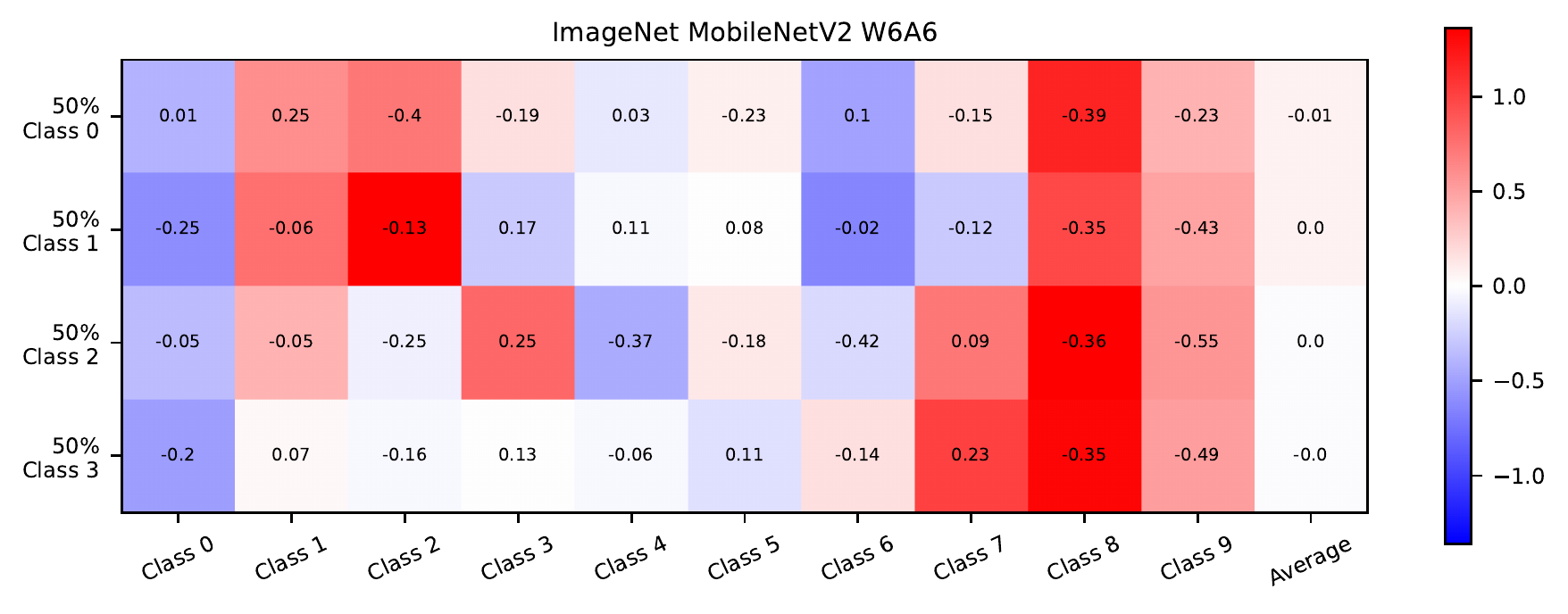}
\end{figure}
\vspace{-1cm}
\begin{figure}[H]
    \centering
    \includegraphics[width=\linewidth]{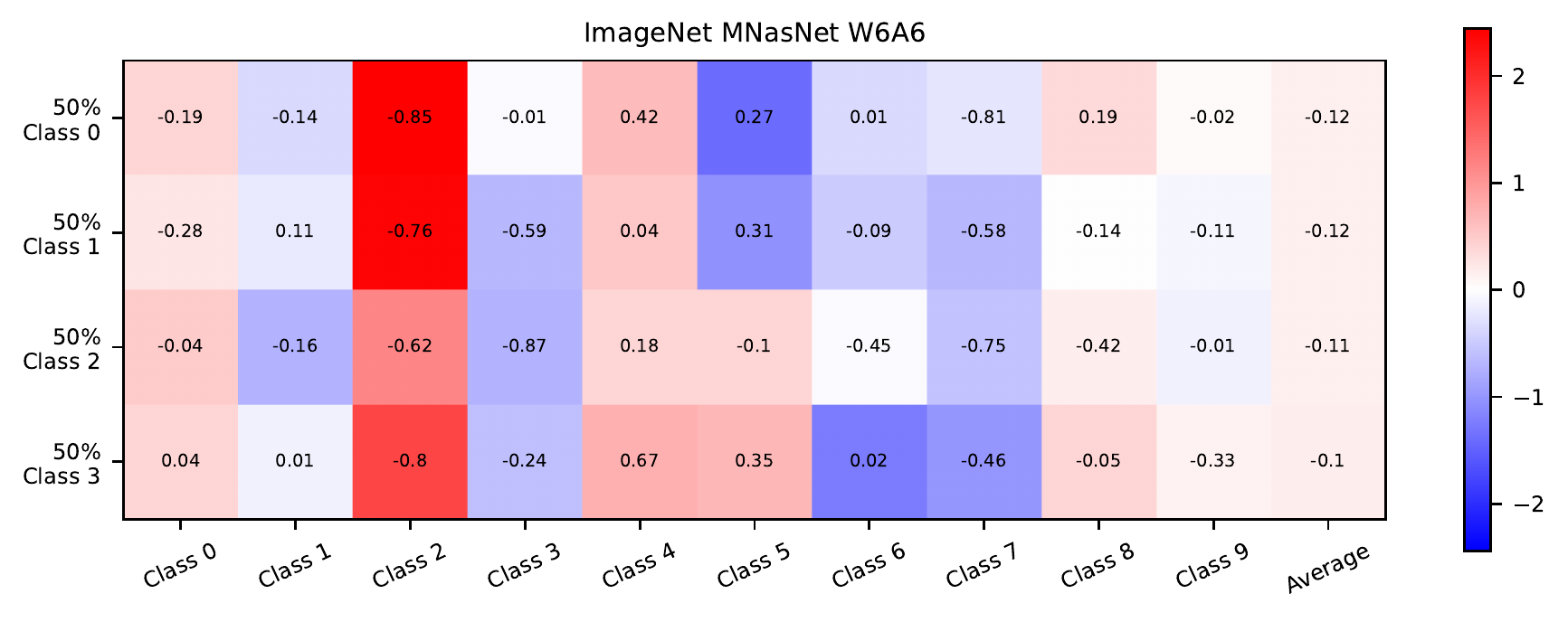}
\end{figure}
\vspace{-1cm}
\begin{figure}[H]
    \centering
    \includegraphics[width=\linewidth]{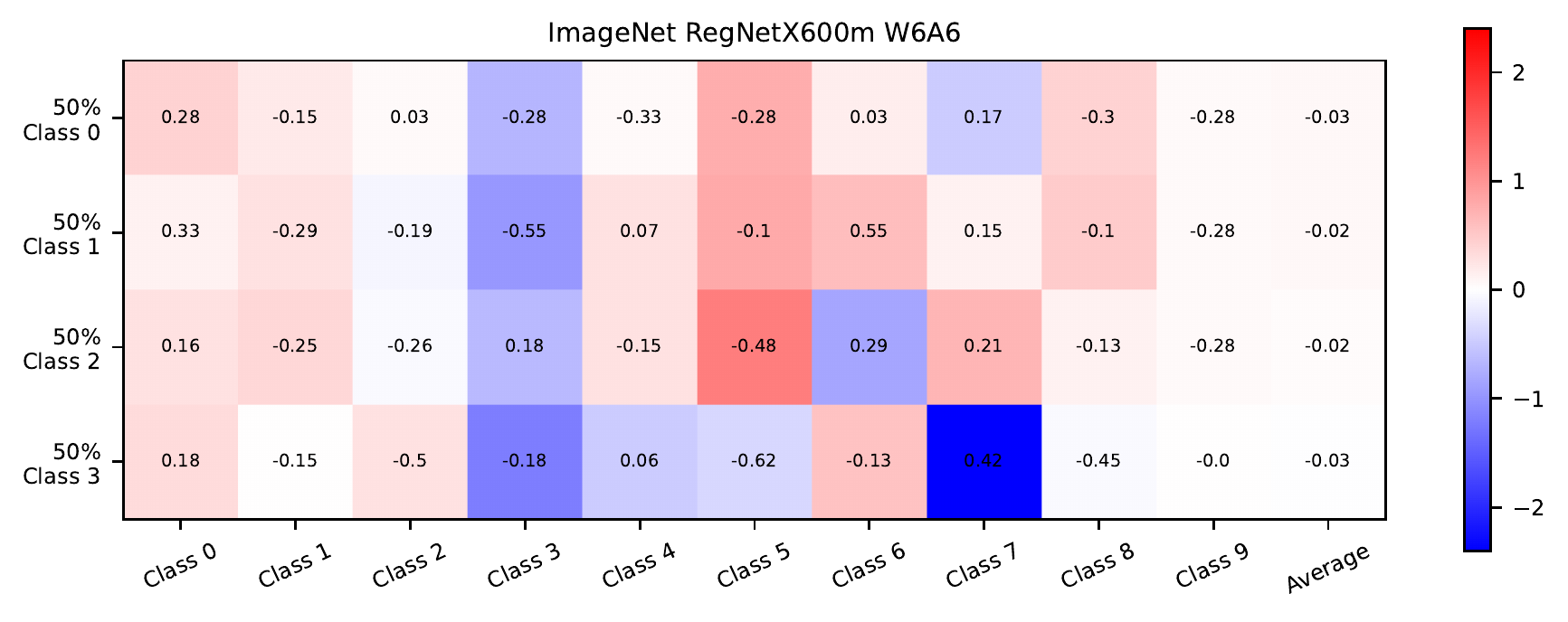}
\end{figure}
\vspace{-1cm}
\begin{figure}[H]
    \centering
    \includegraphics[width=\linewidth]{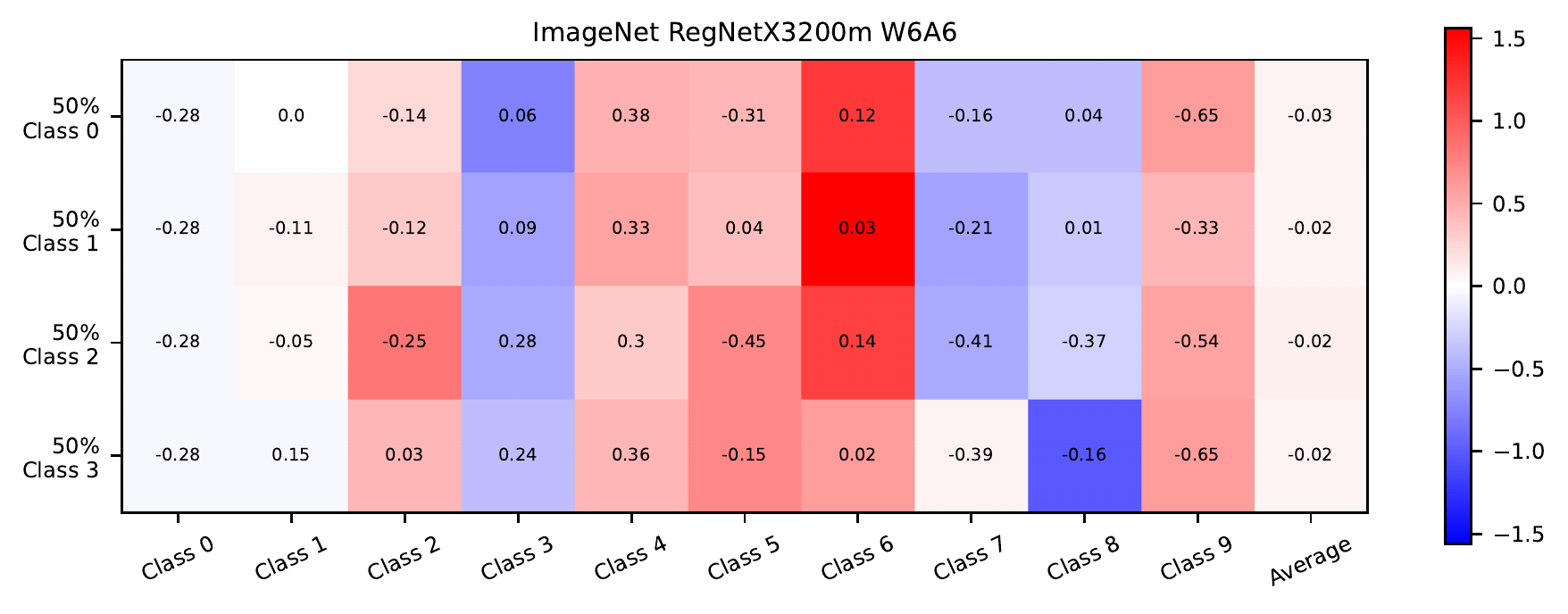}
\end{figure}

\end{appendices}

\end{document}